\documentclass[preprint,12pt,authoryear]{elsarticle}

\usepackage{environ}

\newlength{\completemargin}
\newlength{\widelinewidth}
\NewEnviron{widefigure}[1][]%
{\begin{figure}[#1]
        \edef\side{l}%
        \makebox[\textwidth][\side]{%
            \begin{minipage}[t]{\widelinewidth}
                \BODY
            \end{minipage}%
        }%
    \end{figure}}

\usepackage{amssymb}
\usepackage{amsmath}
\usepackage{bm}
\usepackage{mathrsfs}
\usepackage[hidelinks]{hyperref}
\usepackage{booktabs}
\usepackage[acronym,toc]{glossaries}
\usepackage[table]{xcolor}
\makeglossaries
\newacronym{A2D2}{A2D2}{Audi Autonomous Driving Dataset}
\newacronym{DoF}{DoF}{degrees of freedom}
\newacronym{ICP}{ICP}{Iterative Closest Point}
\newacronym{MLS}{MLS}{Mobile Laser Scanning}
\newacronym{ALS}{ALS}{Airborne Laser Scanning}
\newacronym{ODD}{ODD}{Operational Design Domain}
\newacronym{LOD}{LOD}{Level of Detail}
\newacronym{NIST}{NIST}{National Institute of Standards and Technology}
\newacronym{ADAS}{ADAS}{Advanced Driver Assistance System}
\newacronym{AD}{AD}{Autonomous Driving}
\newacronym{OEM}{OEM}{Original Equipment Manufacturer}
\newacronym{GRPC}{GRPC}{global registration method using a robust phase correlation method}
\newacronym{SuT}{SuT}{System under Test}
\newacronym{ADS}{ADS}{Automated Driving System}
\newacronym{ADF}{ADF}{Automated Driving Function}
\newacronym{IL}{IL}{Imitation Learning}
\newacronym{LLF}{LLF}{low-level fusion}
\newacronym{MLF}{MLF}{mid-level fusion}
\newacronym{HLF}{HLF}{high-level fusion}
\newacronym{FDTD}{FDTD}{Finite-Difference Time-Domain}
\newacronym{FEM}{FEM}{Finite Element Method}
\newacronym{MoM}{MoM}{Method of Moments}
\newacronym{BSDF}{BSDF}{bidirectional scattering distribution function}
\newacronym{BRDF}{BRDF}{bidirectional reflectance distribution function}
\newacronym{BTDF}{BTDF}{bidirectional transmittance distribution function}
\newacronym{CRS}{CRS}{coordinate reference system}
\newacronym{V2X}{V2X}{Vehicle-to-Everything}
\newacronym{C-V2X}{C-V2X}{Cellular V2X}
\newacronym{LTE-V2X}{LTE-V2X}{LTE-V2X}
\newacronym{5G-V2X}{5G-V2X}{5G-V2X}
\newacronym{6G-V2X}{6G-V2X}{6G-V2X}
\newacronym{ViL}{ViL}{Vehicle-in-the-Loop}
\newacronym{DiL}{DiL}{Driver-in-the-Loop}
\newacronym{VHiL}{VHiL}{Vehicle Hardware-in-the-Loop}
\newacronym{HiL}{HiL}{Hardware-in-the-Loop}
\newacronym{SiL}{SiL}{Software-in-the-Loop}
\newacronym{XiL}{XiL}{Everything-in-the-Loop}
\newacronym{ISO}{ISO}{International Organization for Standardization}
\newacronym{DSRC}{DSRC}{Dedicated short-range communications}
\newacronym{FOV}{FOV}{field of view}
\newacronym{HMI}{HMI}{Human Machine Interaction}
\newacronym{SLAM}{SLAM}{Simultaneous Localization And Mapping}
\newacronym{NDT}{NDT}{Normal Distribution Transform}
\newacronym{GDF}{GDF}{Geographic Data File}
\newacronym{NDS}{NDS}{Navigation Data Standard}
\newacronym{XML}{XML}{Extensible Markup Language}
\newacronym{GML}{GML}{Geography Markup Language}
\newacronym{OSI}{OSI}{Open Simulation Interface}
\newacronym{OSM}{OSM}{OpenStreetMap}
\newacronym{OGC}{OGC}{Open Geospatial Consortium}
\newacronym{UTM}{UTM}{Universal Transverse Mercator}
\newacronym{EPSG}{EPSG}{European Petroleum Survey Group}
\newacronym{ETRS89}{ETRS89}{European Terrestrial Reference System 1989}
\newacronym{DHHN2016}{DHHN2016}{Deutsche Haupthöhennetz 2016}
\newacronym{ADE}{ADE}{Application Domain Extension}
\newacronym{RTK}{RTK}{Real Time Kinematic}
\newacronym{ROS}{ROS}{Robot Operating System}
\newacronym{FME}{FME}{Feature Manipulation Engine}
\newacronym{VTD}{VTD}{Virtual Test Drive}
\newacronym{ETL}{ETL}{extract, transform, load}
\newacronym{CDR}{CDR}{Common Data Representation}
\newacronym{ORM}{ORM}{Object Relational Mapper}
\newacronym{UML}{UML}{Unified Modeling Language}
\newacronym{AV}{AV}{automated vehicle}
\newacronym{ACES}{ACES}{autonomous-connected-electric-shared}
\newacronym{UNECE}{UNECE}{United Nations Economic Commission for Europe}
\newacronym{DEM}{DEM}{Digital Elevation Model}
\newacronym{RMSE}{RMSE}{root mean square error}
\newacronym{UAV}{UAV}{Unmanned Aerial Vehicle}
\newacronym{t-SNE}{t-SNE}{t-distributed stochastic neighbor embedding}

\usepackage[binary-units]{siunitx}
\DeclareSIUnit\px{px}
\DeclareSIUnit{\million}{M}
\usepackage{placeins}
\usepackage{xcolor}
\definecolor{tum-orange}		{cmyk}{0.00,0.65,0.95,0.00}
\usepackage{eso-pic}
\usepackage{tikz}

\newcommand{\preprintnotice}{%
  \begin{tikzpicture}[remember picture, overlay]
    \node[
      anchor=north,
      yshift=-1.2cm,
      text width=0.85\paperwidth,
      fill=tum-orange!10,
      draw=tum-orange,
      line width=0.5pt,
      rounded corners=1pt,
      inner sep=6pt,
      font=\small
    ] at (current page.north) {
			\normalfont
      \href{https://doi.org/10.1016/j.jag.2026.105533}{This paper has been published in substantially revised form as: Schwab, B., Kolbe, T.H., 2026. Radiometric fingerprinting of object surfaces using mobile laser scanning and semantic 3D road space models. \textit{International Journal of Applied Earth Observation and Geoinformation} 153, 105533. doi:10.1016/j.jag.2026.105533}. \\
			\vspace*{0.5\baselineskip}
			\textbf{The published article is open access and available at \url{https://doi.org/10.1016/j.jag.2026.105533}}.
    };
  \end{tikzpicture}%
}
\AddToShipoutPictureBG*{\preprintnotice}

\journal{International Journal of Applied Earth Observation and Geoinformation}

\begin{document}

\begin{frontmatter}


	\author[label1]{Benedikt Schwab\corref{cor1}}
	\ead{benedikt.schwab@tum.de}

	\author[label1]{Thomas H. Kolbe} 

	\affiliation[label1]{organization={Chair of Geoinformatics, TUM School of Engineering and Design, Technical University of Munich},
		addressline={Arcisstraße 21},
		city={Munich},
		postcode={80333},
		country={Germany}}

	\title{Radiometric fingerprinting of object surfaces using mobile laser scanning and semantic 3D road space models}


	\begin{abstract}
		Although semantic 3D city models are internationally available and becoming increasingly detailed, the incorporation of material information remains largely untapped.
		However, a structured representation of materials and their physical properties could substantially broaden the application spectrum and analytical capabilities for urban digital twins.
		At the same time, the growing number of repeated mobile laser scans of cities and their street spaces yields a wealth of observations influenced by the material characteristics of the corresponding surfaces.
		To leverage this information, we propose radiometric fingerprints of object surfaces by grouping LiDAR observations reflected from the same semantic object under varying distances, incident angles, environmental conditions, sensors, and scanning campaigns.
		Our study demonstrates how \SI{312.4} million individual beams acquired across four campaigns using five LiDAR sensors on the \gls{A2D2} vehicle can be automatically associated with \num{6368} individual objects of the semantic 3D city model.
		The model comprises a comprehensive and semantic representation of four inner-city streets at \gls{LOD} 3 with centimeter-level accuracy.
		It is based on the CityGML 3.0 standard and enables fine-grained sub-differentiation of objects.
		The extracted radiometric fingerprints for object surfaces reveal recurring intra-class patterns that indicate class-dominant materials.
		The semantic model, the method implementations, and the developed geodatabase solution \emph{3DSensorDB} are released under: \url{https://github.com/tum-gis/sensordb}
	\end{abstract}



	\begin{keyword}


		Radiometric fingerprint
		\sep Mobile laser scanning
		\sep Semantic road space model
		\sep CityGML
		\sep 3DSensorDB
	\end{keyword}

\end{frontmatter}



\section{Introduction}
Driven by a broad range of urban digital twin applications and analytics capabilities, semantic 3D city models are becoming increasingly detailed.
While \gls{LOD}2 building models are already maintained and provided by public authorities for entire countries, such as Germany, Poland, and most of Japan, a growing number of cities are now capturing their road spaces and increasingly demanding \gls{LOD}3 building models \citep{wysockiReviewingOpenData2024}.
However, the representation and reconstruction of surface materials and their physical properties remain underexplored, despite their relevance for a wide range of applications.
These include building energy demand estimation \citep{harterLifeCycleAssessment2023}, microclimate simulations \citep{chenCombiningCityGMLFiles2020}, solar potential analysis \citep{xuSolarCitiesMultiplereflection2025}, sensor simulations \citep{winiwarterVirtualLaserScanning2022,lopezEnhancingLiDARPoint2025}, and predictive maintenance through the detection of traffic signs with degraded reflectivity.

\begin{figure}[htb]
    \centering
    \includegraphics[width=\linewidth]{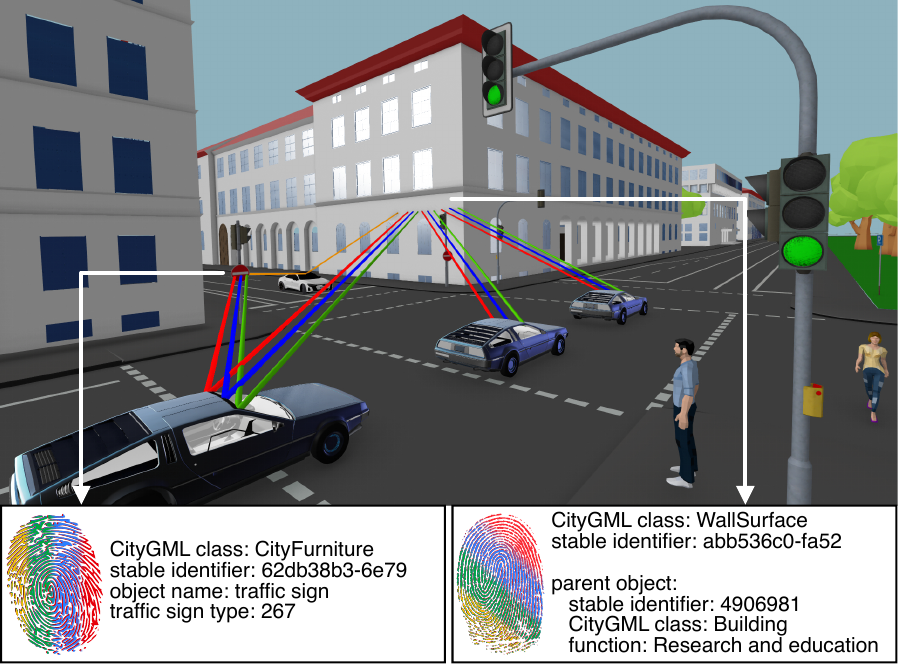}
    \caption[Deriving radiometric fingerprints]{Deriving radiometric fingerprints by grouping electromagnetic radiation responses of object surfaces measured by multiple LiDAR sensors across varying distances, angles, and environmental conditions to infer surface-specific properties and material composition information.\protect\footnotemark}
    \label{fig:radiometric-fingerprint-concept}
\end{figure}

Concurrently, advances in autonomous driving and robotics are accelerating the large-scale deployment of sensors that repeatedly scan road spaces across varying distances, viewing angles, and environmental conditions.
This raises the question of how the acquired sensor data can be utilized to obtain insights about the material properties of object surfaces.
To address this specifically for LiDAR sensors, we introduce the concept of radiometric fingerprinting, as illustrated in \autoref{fig:radiometric-fingerprint-concept}.
A radiometric fingerprint denotes the characteristic and repeatable response of an object's surface to electromagnetic radiation as a function of measurement conditions, such as the range, incident angle, wetness, and atmospheric transmittance.
Radiometric fingerprints may either directly serve applications, for example by providing reflectance characteristics of surfaces, or enable the assignment of physical material properties by matching them to entries from a material library.

\addtocounter{footnote}{-1}

Analyzing the sensor response characteristics of individual object surfaces across multiple sensors under varying environmental conditions requires associating each sensor observation with consistently identifiable objects.
Despite promising developments, semantic 3D city models with comprehensive representations of road spaces at \gls{LOD}3 have not yet been leveraged for such analyses.
The main contributions of this paper are as follows:
\begin{itemize}
    \item Proposing a method for associating individual LiDAR beams with the corresponding object surfaces from a semantic 3D road space model and extracting radiometric fingerprints that characterize their material-specific reflective properties.
    \item Creating the first semantic road space model at \gls{LOD}3 in CityGML 3.0 with \num{15816} individual objects as open data. The semantic model, published as open data, offers  comprehensive coverage of four inner-city streets in Ingolstadt, Germany, spanning an area of \SI[parse-numbers=false]{219\times427}{\meter}, with relative accuracy in the centimeter range.
    \item Conducting four laser scanning campaigns in this study area using an established vehicle equipped with five LiDAR sensors, capturing repeated observations of object surfaces across varying ranges, zenith and azimuth angles, sensors, and campaign conditions.
          \footnotetext[\thefootnote]{\href{https://skfb.ly/owytv}{Delorean} by Luis Adrian licensed under \href{http://creativecommons.org/licenses/by/4.0/}{CC-BY-4.0}.}
    \item Introducing the \emph{3DSensorDB}, a novel database solution that complements the functionality of \emph{3DCityDB}\footnote{\url{https://www.3dcitydb.org}} to associate georeferenced LiDAR point clouds with semantic models.
\end{itemize}

\section{Background and related work}

\subsection{Interoperable and semantic modeling of road space}
To exchange and manage semantic 3D city models in an interoperable way, the standard CityGML has been established internationally.
The conceptual data model of CityGML version 3.0 was published by the \gls{OGC} in \citeyear{kolbeOGCCityGeography2021} \citep{kolbeOGCCityGeography2021}, while the corresponding GML encoding specification followed in \citeyear{kutznerOGCCityGeography2023} \citep{kutznerOGCCityGeography2023}.
The new version features a thoroughly revised Transportation module that supports a wide range of mobility-related use cases, including traffic simulation and multimodal navigation \citep{beilApplicationsSemantic3D2024}.
Moreover, CityGML supports the representation of appearances that describe observable surface properties, such as RGB images, infrared radiation or noise pollution.
\gls{LOD}2 building models covering entire countries, such as Germany and The Netherlands, are maintained and provided by public authorities with absolute centimeter-level accuracy and stable identifiers.
More than 215 million building models are available as open data across countries, including Japan, Poland, Germany, The Netherlands, and the United States \citep{wysockiReviewingOpenData2024}.

The standard OpenDRIVE was initially developed to exchange road network information for vehicle dynamics simulations and is now also used in the context of automated driving systems development.
The current version 1.8.1 was released by the standardization organization \cite{asamOpenDRIVEV181User2024}.
OpenDRIVE is typically created on a per-project basis and is supported by driving and traffic simulators.
It defines its own geometry model, whereas all objects are defined relative to the road's reference line with parametric geometries.
OpenDRIVE's nested parametric geometry model hinders the generation of spatial indexes.
Nevertheless, efficient spatial analysis can be achieved by converting OpenDRIVE datasets to CityGML \citep{schwabSpatiosemanticRoadSpace2020}.

\subsection{LiDAR datasets}
Over the past year, numerous \gls{MLS} datasets of road environments have been released as open data with particular emphasis on autonomous driving and robotics.
They are typically tailored to support perception research, including tasks like object tracking, classification, and semantic segmentation.
Notable examples include \emph{KITTI} \citep{geigerAreWeReady2012}, \emph{SemanticKITTI} \citep{behleySemanticKITTIDatasetSemantic2019}, and the \acrfull{A2D2} \citep{geyerA2D2AudiAutonomous2020}.
However, they commonly provide annotated sensor data without corresponding semantic models of the comprehensive road environment.
Although the urban digital twin dataset TUM2TWIN provides semantic 3D models and post-processed point clouds from mobile mapping platforms, it does not contain raw sensor measurements, bus signals, and configurations of an automated driving vehicle \citep{wysockiTUM2TWINIntroducingLargeScale2025}.

\subsection{Physical principles of laser scanning intensity}
LiDAR intensity measurements are influenced by parameters related to the sensor, target, and atmospheric conditions \citep{hofleCorrectionLaserScanning2007,wagnerGaussianDecompositionCalibration2006}.
The received signal power $P_r$ can be described by the radar range equation \citep{jelalianLaserRadarSystems1992}:
\begin{equation}
    P_r = \frac{P_t D_r^2}{4\pi R^4 \beta_t^2} \eta_{\text{sys}} \eta_{\text{atm}} \sigma,
    \label{eq:radar-range}
\end{equation}
where:
\begin{itemize}
    \itemsep0em
    \item $P_t$: transmitted signal power (\unit{\watt})
    \item $D_r$: diameter of the receiver aperture (\unit{\meter})
    \item $R$: range from sensor to target (\unit{\meter})
    \item $\beta_t$: laser beamwidth (\unit{rad})
    \item $\eta_{\text{sys}}$: system transmission factor
    \item $\eta_{\text{atm}}$: atmospheric transmission factor
    \item $\sigma$: target cross section (\unit{\meter^2})
\end{itemize}
The received signal power is influenced by the target's cross section, which is defined as:
\begin{equation}
    \sigma = \frac{4\pi}{\Omega} \rho A_s,
    \label{eq:cross-section}
\end{equation}
where:
\begin{itemize}
    \itemsep0em
    \item $\Omega$: solid angle of scattering (\unit{\steradian})
    \item $\rho$: target reflectance
    \item $A_s$: target surface area illuminated by the laser (\unit{\meter^2})
\end{itemize}
Reflectance ($\rho$) describes the ratio of reflected to incident radiation and depends on the laser wavelength and the directions of the incident and reflected beams.
Materials exhibit varying degrees of specular and diffuse reflection, which influences how beams are scattered from their surfaces.

\subsection{Bidirectional reflectance distribution functions}

In reflectance theory, the \gls{BRDF} formalizes the reflectance behavior of a target surface  \citep{nicodemusGeometricalConsiderationsNomenclature1977}.
The \gls{BRDF} $f$, expressed in inverse steradians, is defined as
\begin{equation}
    f(\theta_i, \phi_i; \theta_s, \phi_s; \lambda) = \frac{L_s(\theta_i, \phi_i; \theta_s, \phi_s; \lambda)}{E_i(\theta_i, \phi_i; \lambda)},
    \label{eq:brdf_lambda}
\end{equation}
where $\lambda$ denotes the wavelength, $L_s$ is the scattered radiance, and $E_i$ is the incident irradiance.
The angles $\theta$ and $\phi$ represent the zenith and azimuth directions, respectively.
Since LiDAR sensors are typically monostatic, which means that the transmitter and receiver are colocated, the incident and backscattered beams can be assumed to be colinear, such that $\theta_i \approx \theta_s$ and $\phi_i \approx \phi_s$.
Although surface reflectance is wavelength-dependent, LiDAR sensors generally operate at a fixed wavelength, such as \SI{532}{\nano\meter} (bathymetric), \SI{905}{\nano\meter}, \SI{1064}{\nano\meter}, and \SI{1550}{\nano\meter}.

Many \glspl{BRDF} have been proposed that differ in terms of modeled material effects, controllability, computational complexity, and acquisition method \citep{guarneraBRDFRepresentationAcquisition2016}.
Phenemenological \glspl{BRDF} approximate real-world reflectance characteristics using analytical formulas, such as the Phong and Blinn-Phong \glspl{BRDF}.
In contrast, physically-based \glspl{BRDF} are derived from principles of physics and optics modeling light-matter interactions using measurable physical quantities.
Examples include the Cook-Torrance and Oren-Nayar models.
For data-driven \glspl{BRDF}, reflectance characteristics are measured empirically, stored in lookup tables and retrieved using interpolation techniques.
To reduce storage requirements or improve editability, these measured \glspl{BRDF} are often approximated using suitable function spaces \citep{guarneraBRDFRepresentationAcquisition2016}.

Anisotropic \glspl{BRDF} can represent materials whose appearance varies with the orientation around the surface normal.
Examples include brushed metal and hair, where the directional structure significantly influences the reflected light.
In contrast, isotropic \glspl{BRDF} are invariant to surface orientation and therefore do not depend on the azimuthal angles $\phi_i$ and $\phi_s$.

\subsection{Material property databases}

\cite{dupuyAdaptiveParameterizationEfficient2018} scanned 62 materials over a wavelength range of \SIrange{360}{1000}{\nano\meter}, with a spectral resolution of approximately \SI{4}{\nano\meter}, using a goniophotometer with an adaptive scanning parametrization that automatically prioritizes important regions.
\cite{meerdinkECOSTRESSSpectralLibrary2019} collected over 3400 spectra of natural and artificial materials.
Although the dataset spans a broad wavelength range of \SIrange{0.35}{15.4}{\micro\meter}, it does not account for variations in incident or reflection direction.
\cite{polyanskiyRefractiveindexinfoDatabaseOptical2024} provides a refractive index database of 605 materials compiled from peer-reviewed publications, manufacturers' datasheets, and authoritative reference texts.

\subsection{LiDAR simulators}
Early LiDAR simulators approximated the time-of-flight principle by casting single rays to model beam direction, occlusion, and divergence.
These simulators typically relied on ray-casting techniques to generate visually plausible point clouds \citep{gschwandtnerBlenSorBlenderSensor2011}.
More advanced simulators model each LiDAR pulse using multiple rays to approximate a uniform or Gaussian energy distribution and detect one or multiple returns \citep{winiwarterVirtualLaserScanning2022,lopezEnhancingLiDARPoint2025}.
By recording the complete waveform of the backscattered signal, full-waveform LiDAR sensors provide more detail about illuminated objects compared to discrete-return systems.
Well-known full-waveform simulators include DART \citep{yangComprehensiveLiDARSimulation2022} and HELIOS++ \citep{winiwarterVirtualLaserScanning2022}, which supports simulations of terrestrial, mobile, UAV-borne, and airborne laser scanning campaigns.

\subsection{Radiometric correction and calibration}
The raw intensity values provided by real-world LiDAR sensors indicate the amount of scattered radiance, but not in physically well-defined radiometric quantities.
Sensor manufacturers typically scale and adjust the measured radiance to non-standardized value ranges without disclosing the applied functions and parameters.
\cite{kashaniReviewLIDARRadiometric2015} distinguish four levels of LiDAR intensity processing.
Raw intensity values are obtained directly from the sensor without modification (Level 0).
Intensity correction (Level 1) aims to reduce or eliminate variations caused by acquisition parameters, such as range and angle of incidence, yielding pseudo-reflectance values.
Intensity normalization (Level 2) involves scaling and shifting to align overlapping scan strips and achieve overall consistency.
Rigorous radiometric correction and calibration (Level 3) entails assessing the sensor's response to targets with known reflectance properties to derive calibration functions.
When meticulously applied in combination with Level 1 corrections, this process enables the retrieval of true surface reflectance values.

\citeauthor{laaschAutomaticInsituRadiometric2025} proposed a method for automatic in-situ radiometric calibration for \gls{MLS} that compensates for distance and angle of incidence effects without relying on overlapping scans or discrete scan positions.
Their local correction strategy includes a semantic segmentation used as a proxy for the surface material \citep{laaschAutomaticInsituRadiometric2025}.

\subsection{Research gap and questions}
While available and increasingly detailed semantic 3D city models serve as an information basis for analytics and applications, they typically lack detailed physical material properties at the level of individual object surfaces.
However, many applications require material information and benefit from more detailed representations, including LiDAR sensor simulation \citep{winiwarterVirtualLaserScanning2022,lopezEnhancingLiDARPoint2025}, building energy demand estimation \citep{harterLifeCycleAssessment2023}, solar potential analysis \citep{xuSolarCitiesMultiplereflection2025}, microclimate simulations \citep{chenCombiningCityGMLFiles2020}, and noise propagation simulations \citep{kumarHarmonizedDataModel2020}.
Furthermore, an increasing number of sensor systems, especially in automated vehicles, repeatedly capture the environment and its objects from diverse angles, distances, at different times and conditions.

To address this gap, we raise the following research questions: First, how can individual LiDAR sensor measurements from multiple campaigns of an automated vehicle can be systematically grouped by the reflected object surface using a detailed semantic 3D city model?
Second, to what extent do quantifiable differences in the radiometric sensor responses of individual object surfaces reveal material-specific characteristics at both the individual and semantically grouped object levels?

\section{Methodology}
\begin{figure}[htb]
    \centering
    \includegraphics[width=\linewidth]{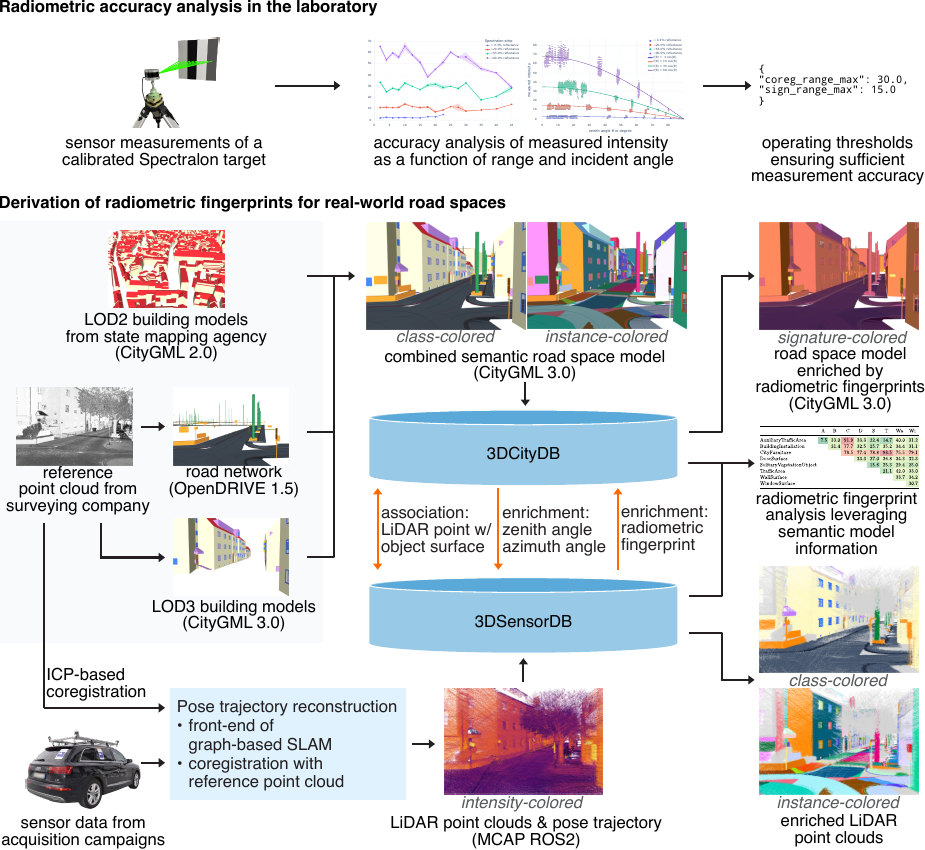}
    \caption{The workflow of this study.}
    \label{fig:study-workflow}
\end{figure}

The workflow of the study is illustrated in \autoref{fig:study-workflow}.
We begin with controlled laboratory experiments to evaluate the radiometric accuracy of the selected LiDAR sensor. The resulting accuracy thresholds are subsequently applied to derive radiometric fingerprints in real-world road environments.

\subsection{Radiometric accuracy analysis in the laboratory}
Due to its widespread use in the fields of automated driving and robotics, we select the Velodyne VLP-16 LiDAR sensor.
The sensor provides sensor-relative range measurements with nanosecond-precision timestamps, which enables the synchronization of multiple sensors.
According to the manufacturer, the measured intensity values are factory-calibrated against a reflectivity reference target calibrated by the \gls{NIST} and remain independent of distance \citep{velodynelidarinc.VLP16UserManual2022}.
\begin{figure}[htb]
    \centering
    \includegraphics[width=.6\linewidth]{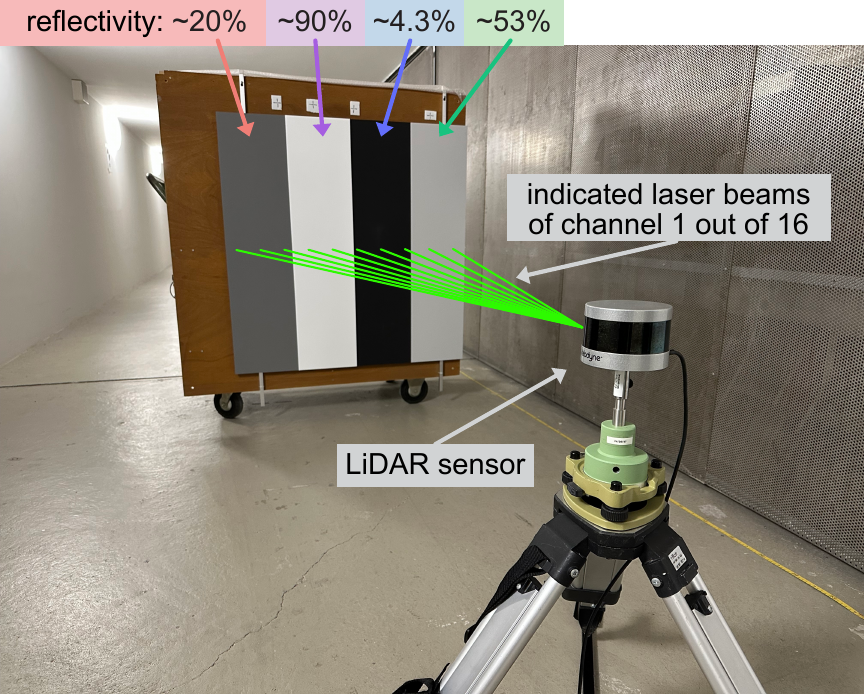}
    \caption{Controlled intensity measurements were conducted with the Velodyne VLP-16 LiDAR sensor and a Spectralon target featuring four calibrated diffuse reflectance strips of approximately \num{20}, \num{90}, \num{4.3}, and \SI{53}{\percent} for varying distances and angles of incidence.}
    \label{fig:experiment-setup}
\end{figure}
To assess the radiometric accuracy of the sensor, we measure the intensity as a function of the range $R$ and zenith angle $\theta$ in laboratory experiments, as illustrated in \autoref{fig:experiment-setup}.
A four-strip Spectralon was utilized as the reference target, whereby each strip's material exhibits Lambertian reflection behavior at a calibrated reflectivity level of \num{4.3}, \num{20}, \num{53}, and \SI{90}{\percent}.

\subsection{Semantic road space model in LOD3}
\begin{figure}[htb]
    \centering
    \includegraphics[width=\linewidth]{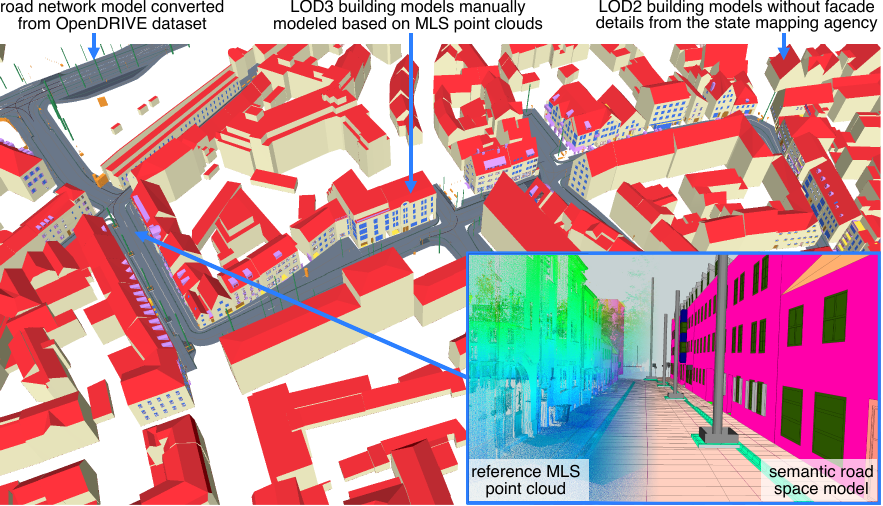}
    \caption{Semantic model of the road spaces in the city of Ingolstadt in Germany according to CityGML 3.0.}
    \label{fig:semantic-environment-model-combined}
\end{figure}

For our study, we select a \SI[parse-numbers=false]{219\times427}{\meter} area in the city center of Ingolstadt, Germany, due to its high urban complexity.
\autoref{fig:semantic-environment-model-combined} shows the respective semantic road space model, which we combine from the state mapping agency and a dedicated surveying campaign by means of their absolute georeferencing.
The semantic model datasets are provided as open data alongside this article.

\subsubsection{Road network and roadside objects}
To acquire detailed information about the road space with reference accuracy at ground level, dedicated surveying campaigns with \gls{MLS} were carried out by the company \emph{3D Mapping Solutions}.
Their high-definition laser scanners mounted on the vehicle can measure up to a million points per second, achieving a distance resolution of \SI{0.1}{\milli\meter} and a point density on the road surface of \SIrange{2}{3}{\milli\meter} in the driven lane \citep{haigermoserRoadTrackIrregularities2015}.
As depicted in \autoref{fig:semantic-environment-model-combined}, the reference \gls{MLS} point clouds serve as a basis for producing the road network datasets with the lane topology and a plethora of roadside objects according to the OpenDRIVE standard.
This includes mapped traffic signs, fences, poles, trees, and controller boxes.
Subsequently, the OpenDRIVE dataset is converted to CityGML 3.0 using the open-source tool r:trån  \citep{schwabSpatiosemanticRoadSpace2020}.
\autoref{fig:road-network-uml} shows the selected classes of the Transportation, CityFurniture, and Vegetation modules of the CityGML 3.0 standard, which are used to represent the road network and roadside objects.

\subsubsection{Building models}
The Bavarian State Mapping Agency offers state-wide \gls{LOD}2 building models as CityGML 2.0 datasets as open data.
The building outlines stem from the official cadastral data with an absolute accuracy of less than \SI{3}{\centi\meter} and the roof shapes are reconstructed from airborne laser scanning point clouds using the method described by \cite{kada3DBuildingReconstruction2009}.
Since a semantic representation of the comprehensive road space is required, the façades are additionally modeled manually concerning their geometry, semantics, and appearance based on the \gls{MLS} point clouds.
We created over 50 \gls{LOD}3 building models with relative accuracy in the range of \SIrange{1}{3}{\centi\meter} and according to the selected CityGML classes depicted in \autoref{fig:building-model-uml}.
As part of this work, we release the CityGML datasets, the editing projects, and a detailed modeling guideline as open data.

\subsection{Sensor data acquisition campaigns and pose trajectory reconstruction}
\begin{figure}[htb]
    \centering
    \includegraphics[width=\linewidth]{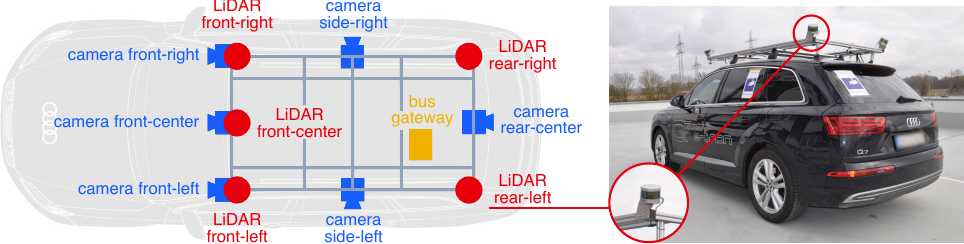}
    \caption{Vehicle of the \acrfull{A2D2} equipped with 6 cameras and 5 LiDAR sensors \citep{geyerA2D2AudiAutonomous2020}.}
    \label{fig:a2d2}
\end{figure}

For the sensor data acquisition campaigns, we use the \gls{A2D2} vehicle, which is equipped with six cameras and five Velodyne VLP-16 LiDAR sensors shown in \autoref{fig:a2d2} \citep{geyerA2D2AudiAutonomous2020}.
We acquired \SI{328.3}{\giga\byte} of real-world sensor data by driving in total four times through the \gls{LOD}3 road space model area.
To obtain comparable sensor measurements under nearly identical environmental conditions, campaigns 2-4 were conducted consecutively on the same day.
Campaign 1 was carried out at a similar time of day and under similar environmental conditions, but during a different season.
\autoref{tab:environmental-conditions} summarizes the environmental conditions for all campaigns.

As with the public \gls{A2D2} dataset, this includes sensor parameters and vehicle bus data, such as linear acceleration, angular velocity, and GPS coordinates.
\begin{table}[htb]
    \sisetup{range-phrase = \text{ to }}
    \footnotesize
    \centering
    \caption{Environmental conditions during the four sensor data acquisition campaigns. Parameter values marked with $\dagger$ are derived from historical meteorological records \citep{timeanddateasWeatherIngolstadtBavaria2025}.}
    \label{tab:environmental-conditions}
    \begin{tabular}{@{} l cccc @{}}
        \toprule
                                                          & Campaign 1                       & \multicolumn{3}{c}{Campaigns 2--4}                                                           \\
        \cmidrule(lr){3-5}
                                                          &                                  & C2                                                   & C3                & C4                \\
        \midrule
        Start time                                        & 13:06:26                         & 12:16:47                                             & 12:27:22          & 12:34:11          \\
        Duration                                          & \SI{90}{\second}                 & \SI{98}{\second}                                     & \SI{107}{\second} & \SI{118}{\second} \\
        Date                                              & 14 January 2020                  & \multicolumn{3}{c}{18 November 2020}                                                         \\
        Astronomical season                               & early winter                     & \multicolumn{3}{c}{late autumn}                                                              \\
        Weather                                           & sunny, few clouds                & \multicolumn{3}{c}{sunny, few clouds}                                                        \\
        Sun exposure                                      & road \& façades shaded           & \multicolumn{3}{c}{road \& façades shaded}                                                   \\
        Precipitation$^\dagger$                           & none                             & \multicolumn{3}{c}{none}                                                                     \\
        Temperature$^\dagger$                             & $\approx$ \SI{7}{\degreeCelsius} & \multicolumn{3}{c}{$\approx$ \SI{6}{\degreeCelsius}}                                         \\
        Humidity$^\dagger$                                & $\approx$ \SI{49}{\percent}      & \multicolumn{3}{c}{$\approx$ \SI{75}{\percent}}                                              \\
        Air pressure$^\dagger$                            & $\approx$ \SI{1017}{\milli\bar}  & \multicolumn{3}{c}{$\approx$ \SI{1027}{\milli\bar}}                                          \\
        First time above \SI{0}{\degreeCelsius}$^\dagger$ & $\approx$ 11{:}00                & \multicolumn{3}{c}{$\approx$ 08{:}00}                                                        \\
        Overnight temp.\ range$^\dagger$                  & \SIrange{-4}{4}{\degreeCelsius}  & \multicolumn{3}{c}{\SIrange{0}{6}{\degreeCelsius}}                                           \\
        Prev.\ day weather$^\dagger$                      & drizzle / fog / interm.\ sun     & \multicolumn{3}{c}{broken clouds / interm.\ sun}                                             \\
        \bottomrule
    \end{tabular}
\end{table}
To utilize existing tools in the robotics domain, we developed an open-source converter from \gls{A2D2} datasets into the bag format of the \gls{ROS} \citep{macenskiRobotOperatingSystem2022}.

Due to the utilization of a standard GPS without error correction, the achieved position accuracy is within the meter range with deviations of up to \SI{9}{\meter}.
Since significantly higher positional and orientation accuracy is required, the vehicle poses are reconstructed during post-processing using exteroceptive sensors.
To attain an accurate and well-controllable pose trajectory reconstruction method, the front-end of the graph-based \gls{SLAM} implementation \emph{Cartographer} is utilized \citep{hessRealtimeLoopClosure2016}.
Based on the trajectories estimated for each submap by the \gls{SLAM} front-end, locally consistent point clouds are derived.
The initial coarse pose estimate of each submap point cloud is determined utilizing the GPS fixes during this time interval.
The submap point cloud is coregistered against the reference \gls{MLS} point cloud of the surveying company.
The coregistration is accomplished using the established point-to-point \gls{ICP} process \citep{masudaRegistrationIntegrationMultiple1996}.

\subsection{3DSensorDB: Scalable management of LiDAR point clouds}
To associate and analyze the georeferenced sensor data in a scalable manner, we developed a novel and freely available database system called 3DSensorDB.
It is based on PostgreSQL and is extended by PostGIS and pgPointcloud.
The 3DSensorDB can operate in standalone mode to manage LiDAR point clouds in a scalable and efficient manner.
Alternatively, it can be operated in conjunction with 3DCityDB V5, which fully supports CityGML 3.0 and enables efficient association, analyses, and enrichments of semantic 3D city models \citep{yaoNew3DCity2025}.
\begin{figure}[htb]
    \centering
    \includegraphics[width=\linewidth]{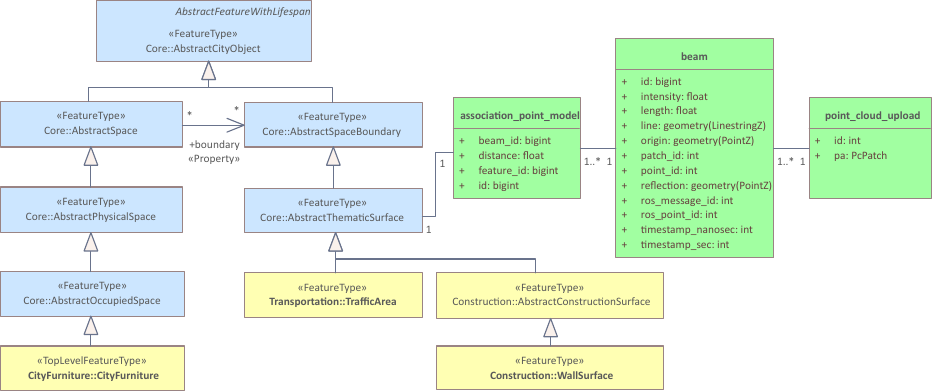}
    \caption{UML diagram of the 3DSensorDB database schema in green alongside a selected set of CityGML classes depicting the association between individual LiDAR beams and their thematic surface from which they were reflected. Due to the mapping to database tables, the 3DSensorDB classes use PostgreSQL data types. Classes of the CityGML 3.0 space concept are colored blue and two classes of thematic modules are shown in yellow as examples.}
    \label{fig:association-uml-diagram}
\end{figure}
The UML diagram of the database schema is illustrated in green in \autoref{fig:association-uml-diagram}.
It enables the registration of sensor platforms, such as vehicles and UAVs, as well as the registration of individual LiDAR sensors.
A campaign comprises the recordings from multiple LiDAR sensors, whereby the acquired point clouds are subdivided into packages.
The point cloud is represented as compressed \texttt{PcPatch} entries of the \emph{pgPointcloud} database extension, along with the envelopes if the sensor positions and measured points.

Our developed companion tool supports reading MCAP-formatted sensor datasets, transforming LiDAR beams from the sensor's coordinate system into the world coordinate system, and partitioning them into uniformly sized point clouds for parallel transfer to the database.
Subsequently, the point cloud packages are decomposed into individual points and inserted into the beam table, which includes dedicated columns to support the association phase.

\subsection{Associating individual LiDAR points with individual object surfaces}
\begin{figure}[htb]
    \centering
    \includegraphics{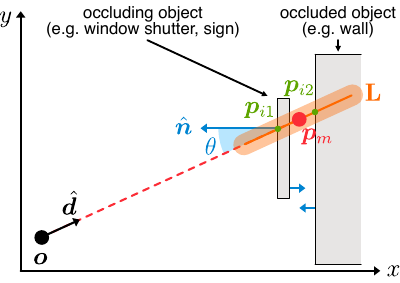}
    \caption{Raycasting-based association approach between LiDAR beam and object surface that is implemented on the 3DSensorDB leveraging the 3DCityDB.}
    \label{fig:association-concept}
\end{figure}

Due to environmental changes, modeling generalizations, and measurement or georeferencing inaccuracies, LiDAR points are typically located within a tolerance range around the surface geometry of the modeled object.
To compensate for such inaccuracies and take occlusions into account, we propose the association approach shown in \autoref{fig:association-concept}.

In a projected, metric, Cartesian coordinate reference system, the sensor position is given by $\bm{o} \in \mathbb{R}^3$ and its orientation by the unit vector $\hat{\bm{d}} \in \mathbb{R}^3$.
Then, the measured reflection point $\bm{p}_m$ corresponding to the range measurement $r \in \mathbb{R}$ is given by:

\begin{equation}
    \bm{p}_m = \bm{o} + r \cdot \hat{\bm{d}}
\end{equation}
To account for measurement uncertainty in the range direction, a line segment $\bm{L}$ of length $\ell$, centered around the measured point $\bm{p}_m$, is considered:
\begin{equation}
    \bm{L} = \left\{
    \bm{p}_m + t \cdot \hat{\bm{d}} \;\middle|\; t \in \left[-\frac{\ell}{2},\, \frac{\ell}{2}\right]
    \right\}
\end{equation}
Let $\mathscr{S}$ denote the set of all surfaces and let each surface $\bm{S} \in \mathscr{S}$ have a corresponding unit normal vector $\hat{\bm{n}}_S$.
The set of candidate surfaces $\mathscr{S}_C$ associated with the measured point is defined as:
\begin{equation}
    \mathscr{S}_C = \left\{ \bm{S} \in \mathscr{S} \;\middle|\;
    \exists\, \bm{p}_S \in \bm{S} :
    \mathrm{dist}\left(\bm{p}_S,\, \mathbf{L}\right) \leq \rho,\quad -\hat{\bm{d}}\cdot\hat{\bm{n}}_{S} \geq 0
    \right\}
\end{equation}
Here, $\bm{p}_S$ represents points located on the surface $S$ and $\mathrm{dist}(\bm{p}_S, \mathbf{L})$ denotes the shortest Euclidean distance from $\bm{p}_S$ to the line segment $\mathbf{L}$.
To approximate the angular measurement uncertainty while preserving computational efficiency, a spherocylinder is employed as the geometric condition.
Its extent is controlled by the threshold parameter $\rho$.

Only for the surface candidates $\mathscr{S}_C$ a computationally expensive 3D intersection is conducted to determine the intersection point $\bm{p}_i = \bm{L} \cap \bm{S}$ between line segment $\bm{L}$ and surface $\bm{S} \in \mathscr{S}_C$.
A signed distance $d_{\text{signed}} \in \mathbb{R}$ from point $\bm{p}_m$ to the intersection point $\bm{p}_i \in \bm{S}$ for each surface candidate $\mathscr{S}_C$ is determined as follows:
\begin{equation}
    d_{\text{signed}}(\bm{p}_m, \bm{p}_i) = (\bm{p}_m - \bm{p}_i) \cdot \hat{\bm{d}}
\end{equation}
The signed distance $d_{\text{signed}}$ enables the candidate surfaces to be ordered according to their relative position along the ray originating from the sensor position $\bm{o}$.
Depending on the use case, the association can be confirmed for all candidates or only a subset.

The zenith angle $\theta \in [0, \frac{\pi}{2}]$ between the incident ray and surface $\bm{S}$ is computed by:
\begin{equation}
    \theta = \arccos\left( -\hat{\bm{d}} \cdot \hat{\bm{n}}_S \right)
\end{equation}
In order to determine the azimuth angle $\phi$, an orthonormal basis $\{\hat{\bm{u}}, \hat{\bm{v}}, \hat{\bm{n}}_S\} \in SO(3)$ is constructed using the Gram-Schmidt process, where $\hat{\bm{u}}$ and $\hat{\bm{v}}$ span the tangent plane of surface $\bm{S}$.
To ensure a consistent orthonormal basis for different rays intersecting with the surface $\bm{S}$, a reference vector $\hat{\bm{a}}$, guaranteed to be non-parallel to the surface normal $\hat{\bm{n}}_S$, is defined as:
\begin{equation}
    \hat{\bm{a}} =
    \begin{cases}
        (1, 0, 0)^\mathsf{T} & \text{if } \sqrt{\hat{\bm{n}}_{S,x}^2 + \hat{\bm{n}}_{S,y}^2} < \epsilon, \\
        (0, 0, 1)^\mathsf{T} & \text{otherwise},
    \end{cases}
\end{equation}
where $\epsilon > 0$ is a small threshold used to avoid numerical instability by detecting near alignment of $\hat{\bm{n}}_S$ with the z-axis.
The perpendicular vectors $\hat{\bm{u}}$ and $\hat{\bm{v}}$ of the orthonormal basis are then determined according to the Gram-Schmidt process:
\begin{eqnarray}
    \hat{\bm{u}} & = & \frac{\hat{\bm{a}} - (\hat{\bm{a}} \cdot \hat{\bm{n}}_S) \hat{\bm{n}}_S}{\| \hat{\bm{a}} - (\hat{\bm{a}} \cdot \hat{\bm{n}}_S) \hat{\bm{n}}_S \|} \\
    \hat{\bm{v}} & = & \hat{\bm{n}}_S \times \hat{\bm{u}}
\end{eqnarray}
To determine the origin of the local orthonormal frame, the sensor position $\bm{o}$ is orthogonally projected onto the plane of the surface $\bm{S}$:
\begin{eqnarray}
    \mathrm{dist}_\perp(\bm{o}, \bm{S}) & = & (\bm{o} - \bm{c}_S) \cdot \hat{\bm{n}}_S, \\
    \bm{o}^\mathrm{proj} & = & \bm{o} - \mathrm{dist}_\perp(\bm{o}, \bm{S}) \hat{\bm{n}}_S,
\end{eqnarray}
where $\bm{c}_S$ denotes the centroid of the surface $\bm{S}$ that is used to compute the orthogonal distance $\mathrm{dist}_\perp(\bm{o}, \bm{S}) \in \mathbb{R}$.
The intersection point $\bm{p}_i$ is transformed into the local orthonormal frame defined by $\{\hat{\bm{u}}, \hat{\bm{v}}, \hat{\bm{n}}_S\}$ and origin $\bm{o}^\mathrm{proj}$:
\begin{equation}
    \bm{p}'_i =
    \begin{pmatrix}
        p'_{i,u} \\
        p'_{i,v} \\
        p'_{i,n}
    \end{pmatrix}
    =
    \begin{pmatrix}
        \hat{\bm{u}}^\top \\
        \hat{\bm{v}}^\top \\
        \hat{\bm{n}}_S^\top
    \end{pmatrix}
    \left( \bm{p}_i - \bm{o}^{\mathrm{proj}} \right),
\end{equation}
where $p'_{i,n} = 0$ by construction since $\bm{p}_i$ lies in the plane.
The normalized azimuth angle $\phi \in [0, 2\pi)$ is then computed as:
\begin{equation}
    \phi = \mathrm{mod}(2\pi + \mathrm{atan2}(p'_{i,v}, p'_{i,u}), 2\pi)
\end{equation}
This approach only associates LiDAR points for which corresponding surface geometries of modeled objects exist.
As the \gls{LOD}3 road space model includes only static elements, points reflected from dynamic objects such as vehicles and pedestrians are not associated.

\subsection{Bidirectional enrichment between LiDAR points and city objects}
Each city object is enriched with statistical summaries of the associated points, including the mean, median, and first and third quartiles of the measured intensity.
These statistics also include distance-based measures, such as the signed distance $d_{\text{signed}}(\bm{p}_m, \bm{p}_i)$ and the minimum distance between the measured point $\bm{p}_m$ and its associated surface $\bm{S}$.

Conversely, each point is enriched with information from the associated city object, including its class name, object name, and unique identifier.
In addition, each point is augmented with the determined geometric attributes, such as the zenith angle $\theta$, azimuth angle $\phi$, and distance-based measures.

\section{Results and discussion}

\subsection{Radiometric accuracy evaluation}
\begin{figure}[htb]
    \centering
    \includegraphics[width=\linewidth]{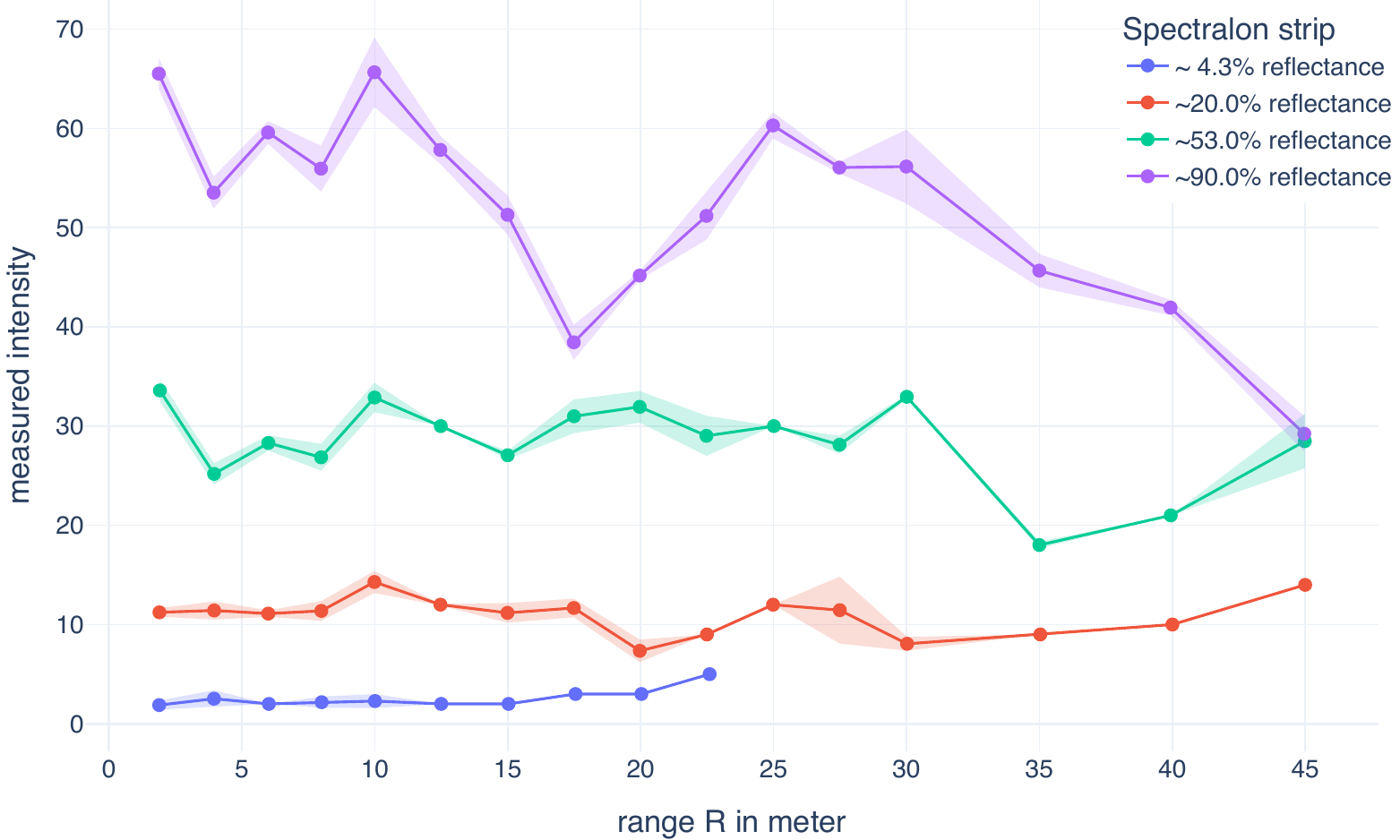}
    \caption{Intensity values measured on the four Spectralon strips as a function of range $R$ with the zenith angle $\theta < 3^{\circ}$, where a point denotes the mean intensity and the shaded region indicates the standard deviation.}
    \label{fig:intensity_mean_distance_mean_laser_id_1}
\end{figure}

\autoref{fig:intensity_mean_distance_mean_laser_id_1} presents the mean intensity and corresponding standard deviation measured on each Spectralon strip in the laboratory as a function of range.
It reveals increased variability at a higher reflectance, especially at \SI{50}{\percent} and \SI{90}{\percent}.
According to the sensor manufacturer, the intensity values between 0 and 100 are supposed to correspond to the reflectivity percentage of a diffuse reflector.
Nonetheless, the experiments confirm a monotonically increasing relation between target reflectivity at ranges up to \SI{40}{\meter}.
However, increased intensity variations are observed at ranges over \SI{30}{\meter} for all remaining Spectralon strips.
Only measurements within \SI{15}{\meter} yield moderately consistent intensity values with acceptable variation.

\begin{figure}[htb]
    \centering
    \includegraphics[width=\linewidth]{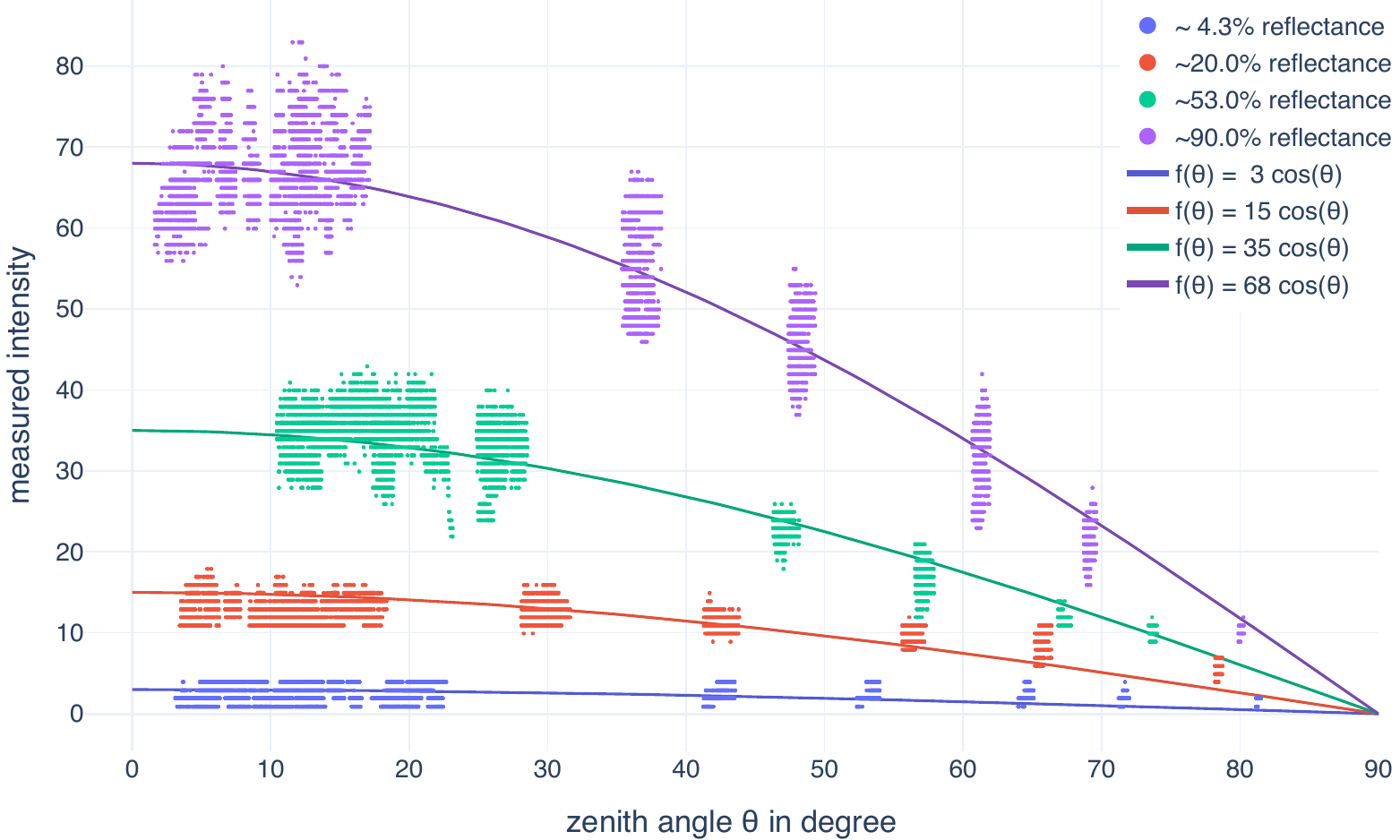}
    \caption{Individual intensity values measured as a function of the zenith angle $\theta$ at a 2-meter distance from the Spectralon rotation center with approximated cosine functions.}
    \label{fig:intensity_incident_angle}
\end{figure}

\autoref{fig:intensity_incident_angle} shows the measured intensity results as a function of the zenith angle.
The observed radiant intensity follows Lambert's cosine law, $I_\theta=I_0 cos(\theta)$, whereas the Spectralon's strips maintain constant radiance regardless of the angle.
Overall, the results demonstrate that the selected sensor is capable of distinguishing between surfaces with noticeably different reflectivities and that intensity measurements are influenced by the zenith angle.

\FloatBarrier
\subsection{Accuracy of the pose trajectory reconstruction and georeferencing}
We conducted the first sensor data acquisition campaign with the \gls{A2D2} vehicle approximately 1.3 years after the reference \gls{MLS} surveying campaign.
Campaigns 2 through 4 followed 5 months later.
To increase the geometric accuracy of pose reconstruction and georeferencing, only points within a range of up to \SI{30}{\meter} were considered.
\begin{figure}[htb]
    \centering
    \includegraphics[width=\linewidth]{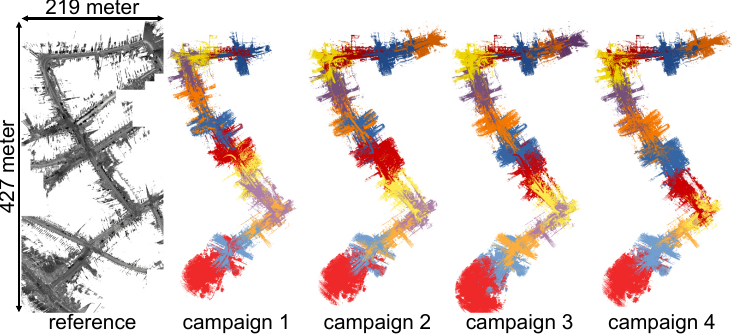}
    \caption{Bird's-eye view of the reference point clouds and the point clouds of each \gls{SLAM} submap from the acquisition campaigns after coregistration. Variations in submap coverage across campaigns necessitate combining the sensor data using each platform's georeferenced pose trajectory, which is also required for associating the sensor data with the semantic road space model.}
    \label{fig:coregistration-submaps}
\end{figure}
\autoref{fig:coregistration-submaps} shows the point clouds of each color-coded \gls{SLAM} submap after successful corregistration with the reference point cloud from a bird's eye view for the \SI[parse-numbers=false]{219\times427}{\meter} area.
As submap boundaries are determined during \gls{SLAM} execution to minimize local drift, their spatial coverage varies across campaigns.

To evaluate the accuracy of the alignment, we use the fitness score $f$, defined as:
\begin{equation}
    f = \frac{\# \text{inlier correspondences}}{\# \text{points in target}},
\end{equation}
where inliers are point correspondences within a distance threshold of \SI{1}{\meter}.
In addition, we compute the \gls{RMSE} between the coregistered submap point cloud and reference point cloud as:
\begin{equation}
    \text{RMSE} = \sqrt{ \frac{1}{N} \sum_{i=1}^{N} \left\| \mathbf{p}_i - \hat{\mathbf{p}}_i \right\|^2 }
\end{equation}
\autoref{tab:fitness_rmse_scores} lists the accuracy scores for each campaign and submap.
\begin{table}[htb]
	\centering
	\footnotesize
	\caption{Fitness and RMSE scores of the point cloud submaps of each campaign after coregistration with the reference point cloud. ET = Elapsed time since reference surveying campaign, SM = submap of graph-based SLAM.}
	\label{tab:fitness_rmse_scores}
	\addtolength{\tabcolsep}{-0.4em}
	\begin{tabular}{@{} l rrrrrrrrrrrrrrrrrrrrrrrrrrrrrrrrrrrr @{}}
		\toprule
		  & \multicolumn{3}{c}{Campaign\,1} & \multicolumn{3}{c}{Campaign\,2} & \multicolumn{3}{c}{Campaign\,3} & \multicolumn{3}{c}{Campaign\,4} \\
		 \cmidrule(lr){2-4} \cmidrule(lr){5-7} \cmidrule(lr){8-10} \cmidrule(lr){11-13} 
		 ET		  & \multicolumn{3}{c}{1\,year 4\,months} & \multicolumn{3}{c}{1\,year 9\,months} & \multicolumn{3}{c}{1\,year 9\,months} & \multicolumn{3}{c}{1\,year 9\,months} \\
		 $\sum$ & \multicolumn{3}{c}{91.1M points} & \multicolumn{3}{c}{109.6M points} & \multicolumn{3}{c}{104.5M points} & \multicolumn{3}{c}{113.2M points} &  \\
		 \cmidrule(lr){2-4} \cmidrule(lr){5-7} \cmidrule(lr){8-10} \cmidrule(lr){11-13} 
		 SM & \multicolumn{1}{c}{\#} & \multicolumn{1}{c}{$f$} & \multicolumn{1}{c}{RMSE} & \multicolumn{1}{c}{\#} & \multicolumn{1}{c}{$f$} & \multicolumn{1}{c}{RMSE} & \multicolumn{1}{c}{\#} & \multicolumn{1}{c}{$f$} & \multicolumn{1}{c}{RMSE} & \multicolumn{1}{c}{\#} & \multicolumn{1}{c}{$f$} & \multicolumn{1}{c}{RMSE} \\
		\midrule
		1 & 7.1M & \cellcolor[rgb]{0.909,0.961,0.798} 96.4\% & \cellcolor[rgb]{0.881,0.949,0.779} 0.18m & 7.3M & \cellcolor[rgb]{0.928,0.969,0.81} 96.2\% & \cellcolor[rgb]{0.971,0.988,0.86} 0.21m & 6.3M & \cellcolor[rgb]{0.63,0.819,0.715} 99.3\% & \cellcolor[rgb]{0.807,0.917,0.76} 0.16m & 6.1M & \cellcolor[rgb]{0.945,0.977,0.824} 95.9\% & \cellcolor[rgb]{0.999,1.0,0.898} 0.22m \\
		2 & 8.2M & \cellcolor[rgb]{0.629,0.816,0.714} 99.3\% & \cellcolor[rgb]{0.622,0.804,0.708} 0.11m & 9.5M & \cellcolor[rgb]{0.648,0.842,0.727} 98.9\% & \cellcolor[rgb]{0.704,0.869,0.741} 0.14m & 6.8M & \cellcolor[rgb]{0.947,0.978,0.828} 95.9\% & \cellcolor[rgb]{0.975,0.99,0.866} 0.21m & 7.4M & \cellcolor[rgb]{0.727,0.881,0.747} 98.3\% & \cellcolor[rgb]{0.681,0.858,0.735} 0.13m \\
		3 & 7.8M & \cellcolor[rgb]{0.638,0.834,0.723} 99.1\% & \cellcolor[rgb]{0.741,0.887,0.751} 0.15m & 9.5M & \cellcolor[rgb]{0.64,0.837,0.725} 99.0\% & \cellcolor[rgb]{0.746,0.89,0.752} 0.15m & 8.0M & \cellcolor[rgb]{0.632,0.822,0.717} 99.2\% & \cellcolor[rgb]{0.635,0.828,0.72} 0.12m & 10.8M & \cellcolor[rgb]{1.0,0.988,0.879} 94.7\% & \cellcolor[rgb]{0.915,0.964,0.802} 0.19m \\
		4 & 7.6M & \cellcolor[rgb]{0.784,0.907,0.758} 97.8\% & \cellcolor[rgb]{0.943,0.976,0.821} 0.20m & 9.4M & \cellcolor[rgb]{0.676,0.856,0.734} 98.7\% & \cellcolor[rgb]{0.847,0.934,0.765} 0.17m & 8.1M & \cellcolor[rgb]{0.635,0.828,0.72} 99.1\% & \cellcolor[rgb]{0.78,0.905,0.757} 0.16m & 25.6M & \cellcolor[rgb]{0.984,0.783,0.711} 92.1\% & \cellcolor[rgb]{1.0,0.993,0.888} 0.23m \\
		5 & 7.1M & \cellcolor[rgb]{0.768,0.9,0.756} 97.9\% & \cellcolor[rgb]{0.881,0.949,0.779} 0.18m & 8.0M & \cellcolor[rgb]{0.815,0.921,0.761} 97.4\% & \cellcolor[rgb]{0.999,0.982,0.869} 0.23m & 7.7M & \cellcolor[rgb]{0.751,0.892,0.753} 98.1\% & \cellcolor[rgb]{0.961,0.984,0.847} 0.21m & 7.9M & \cellcolor[rgb]{0.764,0.898,0.756} 97.9\% & \cellcolor[rgb]{0.851,0.936,0.765} 0.17m \\
		6 & 7.3M & \cellcolor[rgb]{0.662,0.849,0.731} 98.8\% & \cellcolor[rgb]{0.776,0.903,0.757} 0.15m & 8.3M & \cellcolor[rgb]{0.76,0.896,0.755} 98.0\% & \cellcolor[rgb]{0.9,0.957,0.792} 0.19m & 7.0M & \cellcolor[rgb]{0.791,0.91,0.759} 97.7\% & \cellcolor[rgb]{0.982,0.993,0.876} 0.22m & 7.0M & \cellcolor[rgb]{0.807,0.917,0.76} 97.5\% & \cellcolor[rgb]{1.0,0.997,0.895} 0.23m \\
		7 & 7.4M & \cellcolor[rgb]{0.622,0.804,0.708} 99.4\% & \cellcolor[rgb]{0.709,0.871,0.742} 0.14m & 8.3M & \cellcolor[rgb]{0.657,0.846,0.73} 98.8\% & \cellcolor[rgb]{0.858,0.94,0.766} 0.17m & 7.3M & \cellcolor[rgb]{0.643,0.84,0.726} 99.0\% & \cellcolor[rgb]{0.854,0.938,0.766} 0.17m & 6.8M & \cellcolor[rgb]{0.648,0.842,0.727} 98.9\% & \cellcolor[rgb]{0.839,0.931,0.764} 0.17m \\
		8 & 7.0M & \cellcolor[rgb]{0.788,0.909,0.758} 97.7\% & \cellcolor[rgb]{0.999,0.984,0.872} 0.23m & 7.9M & \cellcolor[rgb]{0.69,0.862,0.738} 98.6\% & \cellcolor[rgb]{0.957,0.982,0.84} 0.21m & 7.1M & \cellcolor[rgb]{0.695,0.865,0.739} 98.5\% & \cellcolor[rgb]{0.906,0.96,0.796} 0.19m & 6.8M & \cellcolor[rgb]{0.638,0.834,0.723} 99.0\% & \cellcolor[rgb]{0.819,0.922,0.762} 0.17m \\
		9 & 7.8M & \cellcolor[rgb]{0.875,0.946,0.775} 96.8\% & \cellcolor[rgb]{0.985,0.994,0.879} 0.22m & 7.8M & \cellcolor[rgb]{0.9,0.957,0.792} 96.5\% & \cellcolor[rgb]{0.99,0.827,0.731} 0.29m & 7.0M & \cellcolor[rgb]{0.871,0.945,0.773} 96.8\% & \cellcolor[rgb]{0.995,0.863,0.748} 0.28m & 6.9M & \cellcolor[rgb]{0.819,0.922,0.762} 97.4\% & \cellcolor[rgb]{0.998,0.942,0.81} 0.25m \\
		10 & 8.1M & \cellcolor[rgb]{0.875,0.946,0.775} 96.8\% & \cellcolor[rgb]{0.9,0.957,0.792} 0.19m & 9.3M & \cellcolor[rgb]{0.957,0.982,0.84} 95.7\% & \cellcolor[rgb]{0.961,0.984,0.847} 0.21m & 7.9M & \cellcolor[rgb]{0.909,0.961,0.798} 96.4\% & \cellcolor[rgb]{0.971,0.988,0.86} 0.21m & 7.4M & \cellcolor[rgb]{0.884,0.951,0.781} 96.7\% & \cellcolor[rgb]{0.999,0.972,0.853} 0.24m \\
		11 & 8.4M & \cellcolor[rgb]{0.652,0.844,0.728} 98.9\% & \cellcolor[rgb]{0.723,0.878,0.746} 0.14m & 9.7M & \cellcolor[rgb]{0.815,0.921,0.761} 97.4\% & \cellcolor[rgb]{0.906,0.96,0.796} 0.19m & 8.5M & \cellcolor[rgb]{0.681,0.858,0.735} 98.7\% & \cellcolor[rgb]{0.776,0.903,0.757} 0.15m & 7.9M & \cellcolor[rgb]{0.784,0.907,0.758} 97.8\% & \cellcolor[rgb]{0.862,0.941,0.767} 0.18m \\
		12 & 7.1M & \cellcolor[rgb]{0.957,0.982,0.84} 95.7\% & \cellcolor[rgb]{0.985,0.994,0.879} 0.22m & 9.7M & \cellcolor[rgb]{0.922,0.967,0.806} 96.2\% & \cellcolor[rgb]{0.989,0.996,0.885} 0.22m & 8.4M & \cellcolor[rgb]{0.906,0.96,0.796} 96.4\% & \cellcolor[rgb]{0.952,0.98,0.834} 0.20m & 7.7M & \cellcolor[rgb]{0.887,0.952,0.784} 96.7\% & \cellcolor[rgb]{0.943,0.976,0.821} 0.20m \\
		13 & & &  & 4.8M & \cellcolor[rgb]{0.718,0.876,0.745} 98.3\% & \cellcolor[rgb]{0.868,0.944,0.771} 0.18m & 8.3M & \cellcolor[rgb]{0.9,0.957,0.792} 96.5\% & \cellcolor[rgb]{0.994,0.998,0.892} 0.22m & 4.9M & \cellcolor[rgb]{0.823,0.924,0.762} 97.4\% & \cellcolor[rgb]{0.94,0.975,0.818} 0.20m \\
		14 & & &  & & &  & 6.1M & \cellcolor[rgb]{0.788,0.909,0.758} 97.7\% & \cellcolor[rgb]{0.887,0.952,0.784} 0.18m & & &  \\
		\midrule
		 $\mu$ & 7.6M & \cellcolor[rgb]{0.772,0.902,0.757} 97.9\% & \cellcolor[rgb]{0.865,0.942,0.769} 0.18m & 8.4M & \cellcolor[rgb]{0.791,0.91,0.759} 97.7\% & \cellcolor[rgb]{0.928,0.969,0.81} 0.20m & 7.5M & \cellcolor[rgb]{0.78,0.905,0.757} 97.8\% & \cellcolor[rgb]{0.915,0.964,0.802} 0.19m & 8.7M & \cellcolor[rgb]{0.865,0.942,0.769} 97.0\% & \cellcolor[rgb]{0.937,0.974,0.816} 0.20m \\
		\bottomrule
	\end{tabular}
\end{table}
An average fitness of \SI{97.6}{\percent} and an \gls{RMSE} of \SI{0.19}{\meter} are achieved overall, whereby the most accurate results are delivered by the first campaign, which was conducted with the shortest time interval after the reference survey.
The elevated \gls{RMSE} scores observed for submap 8-10 result from significant scene changes due to facade renovations, including scaffolding and a crane, which were dismantled between the reference survey and the first campaign.
During the fourth campaign, the vehicle was stationary for \SI{33}{\second} due to a blockage caused by another vehicle.
The presence of parked and obstructing vehicles in the immediate vicinity resulted in a comparatively low fitness score of \SI{97.6}{\percent} for submap 4, as well as an increased number of collected points.

\FloatBarrier
\subsection{Semantic road space model in LOD3}
\begin{figure}[htb]
    \centering
    \includegraphics[width=\linewidth]{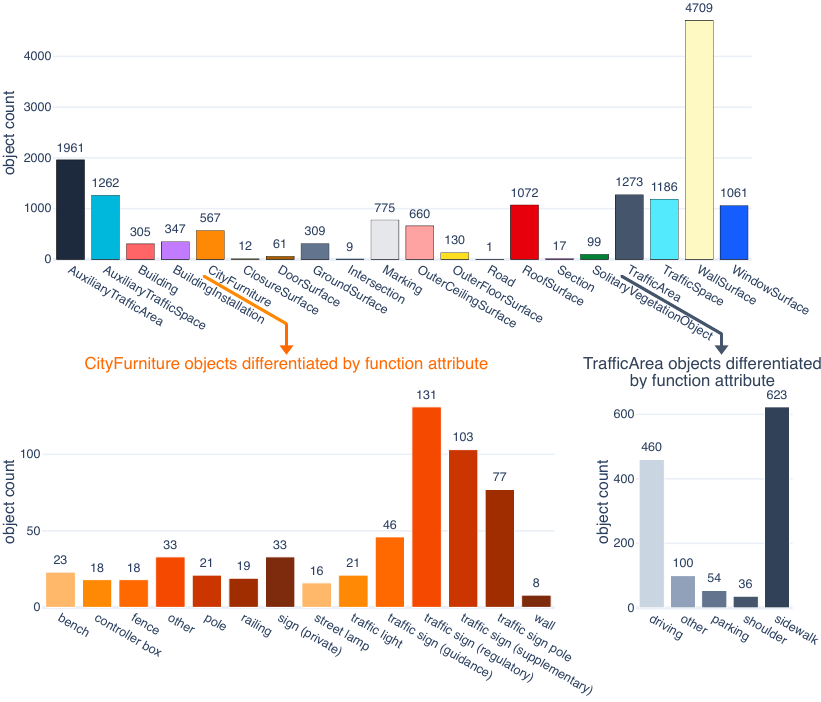}
    \caption{Distribution of the \num{15816} thematic objects in the \gls{LOD}3 road space model broken down by CityGML 3.0 classes (top diagram).
        The model's rich semantic information supports fine subdifferentiations (bottom diagrams) and therefore enables flexible grouping of objects and their associated sensor data.}
    \label{fig:semantic-model-object-distribution}
\end{figure}

The semantic \gls{LOD}3 road space model we created comprises \num{15816} individual thematic objects compliant with the CityGML 3.0 standard.
\autoref{fig:semantic-model-object-distribution} shows the object distribution according to the CityGML classes.
With its comprehensive information, the model supports fine-grained subdifferentiation, such as distinguishing CityFurniture objects by function and identifying traffic signs based on the official catalog.
TrafficArea objects not only include function codes but also encode predecessor-successor relationships.
The structured semantic, geometric, and topological information enables in-depth analysis of the object in the model and its associated sensor data.

\FloatBarrier
\subsection{Association and point cloud enrichment results}
A line segment of length $\ell = \SI{1}{\meter}$ and a spatial threshold of $\rho=\SI{5}{\centi\meter}$ were used to define the spherocylinder condition for the association candidates, whereas the parameters utilized for the association method are summarized in \autoref{tab:automatic-association-parameters} in \ref{appendix:automatic-association}.
Only the first candidate with the lowest signed distance $d_{\text{signed}}$ was confirmed as an association, so that all measured points are associated with an object surface either once or not at all.

\begin{table}[htb]
	\centering
	\footnotesize
	\caption{Distribution of associated points broken down by campaign and range threshold. Points within a 15-meter range are associated more frequently than those beyond this threshold.}
	\label{tab:association-stats}
	\addtolength{\tabcolsep}{-0.15em}
	\begin{tabular}{@{} l rrr rrr rrr @{}}
		\toprule
		& \multicolumn{3}{c}{$r\leq15\,\text{meter}$} & \multicolumn{3}{c}{$r\leq30\,\text{meter}$} & \multicolumn{3}{c}{all ranges}\\
		\cmidrule(lr){2-4}\cmidrule(lr){5-7}\cmidrule(lr){8-10}
		& \multicolumn{2}{c}{associated} & \multicolumn{1}{c}{all} & \multicolumn{2}{c}{associated} & \multicolumn{1}{c}{all} & \multicolumn{2}{c}{associated} & \multicolumn{1}{c}{all}\\
		\cmidrule(lr){2-3}\cmidrule(lr){4-4}\cmidrule(lr){5-6}\cmidrule(lr){7-7}\cmidrule(lr){8-9}\cmidrule(lr){10-10}
		Camp.\,1 & 79.6\% & 58.3M & 73.2M & 75.5\% & 68.8M & 91.1M & 74.6\% & 71.2M & 95.5M \\
		Camp.\,2 & 75.6\% & 65.9M & 87.2M & 70.8\% & 77.6M & 109.6M & 69.8\% & 80.3M & 115.0M \\
		Camp.\,3 & 76.9\% & 66.8M & 86.8M & 72.2\% & 75.5M & 104.5M & 71.3\% & 77.5M & 108.8M \\
		Camp.\,4 & 75.1\% & 72.4M & 96.4M & 71.9\% & 81.4M & 113.2M & 70.9\% & 83.4M & 117.5M \\
		\midrule
		 $\Sigma/\mu$ &76.8\% & 263.5M & 343.6M & 72.6\% & 303.2M & 418.4M & 71.7\% & 312.4M & 436.8M \\
		\bottomrule
	\end{tabular}
\end{table}
As listed in \autoref{tab:association-stats}, out of the 436.8 million points collected across four campaigns, \SI{95.8}{\percent} fall within a range of 30 meters and \SI{60.3}{\percent} lie within 15 meters.
Between \SI{70}{\percent} and \SI{80}{\percent} of the points are associated with object surfaces defined by the semantic road space model.
Unassociated points originate from self-reflections of the vehicle, parked cars, building interiors, and tree canopies, which are not represented in the model, as shown in \autoref{fig:association-comparison} in \ref{appendix:automatic-association}.

\FloatBarrier
\autoref{fig:associated-point-distribution} (top) presents the distribution of associated LiDAR points by 12 classes of CityGML 3.0.
WallSurface and TrafficArea dominate in terms of number of the points due to their extensive spatial coverage in the inner-city environment.
Each class exhibits a roughly uniform distribution across campaigns, due to the consistent route and similar driving speeds through the \gls{LOD}3 area.
\begin{figure}[htb]
    \centering
    \includegraphics[width=\linewidth]{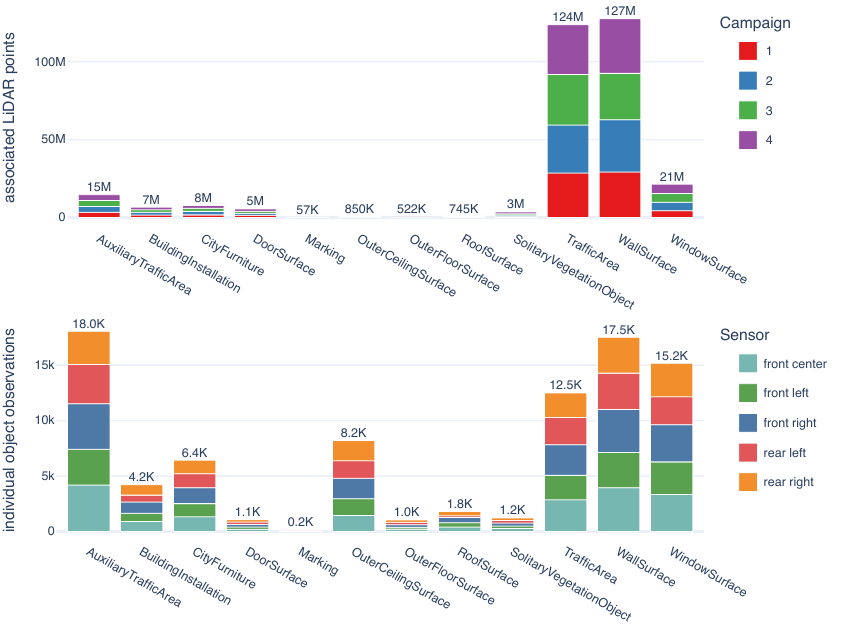}
    \caption{Association distribution of LiDAR points according to the CityGML 3.0 classes (top) and distribution of total object observations (bottom). Our method automatically associated 312.4 million points and identified \num{87345} object observations, each counted individually per sensor and campaign.}
    \label{fig:associated-point-distribution}
\end{figure}
\autoref{fig:associated-point-distribution} (bottom) shows the total number of object observations broken down by object class, where an object observation is counted each time a LiDAR sensor observes an object during a campaign.
Overall, \num{87345} object observations were collected across four campaigns and five sensors.
Classes with high observation counts include WindowSurface, AuxiliaryTrafficArea, OuterCeilingSurface, and CityFurniture, as they comprise a large number of individual objects within the sensors' field of view along the campaign route.

\autoref{fig:street-level-comparisons} depicts the captured sensor data, the semantic model, as well as the enriched sensor data from the same street-level perspective.
The results demonstrate that our proposed method enables the automatic association of point clouds from \gls{MLS} campaigns with medium-priced LiDAR sensors, while reliably identifying even small objects such as window shutters and signs represented in the semantic model.
Based on the semantic model, the zenith and azimuth angles of the incident LiDAR beams on the surface of the object are also calculated, as indicated by the colored point clouds in the bottom row.
\begin{widefigure}[htb]
    \raggedright
    \includegraphics[width=.92\linewidth]{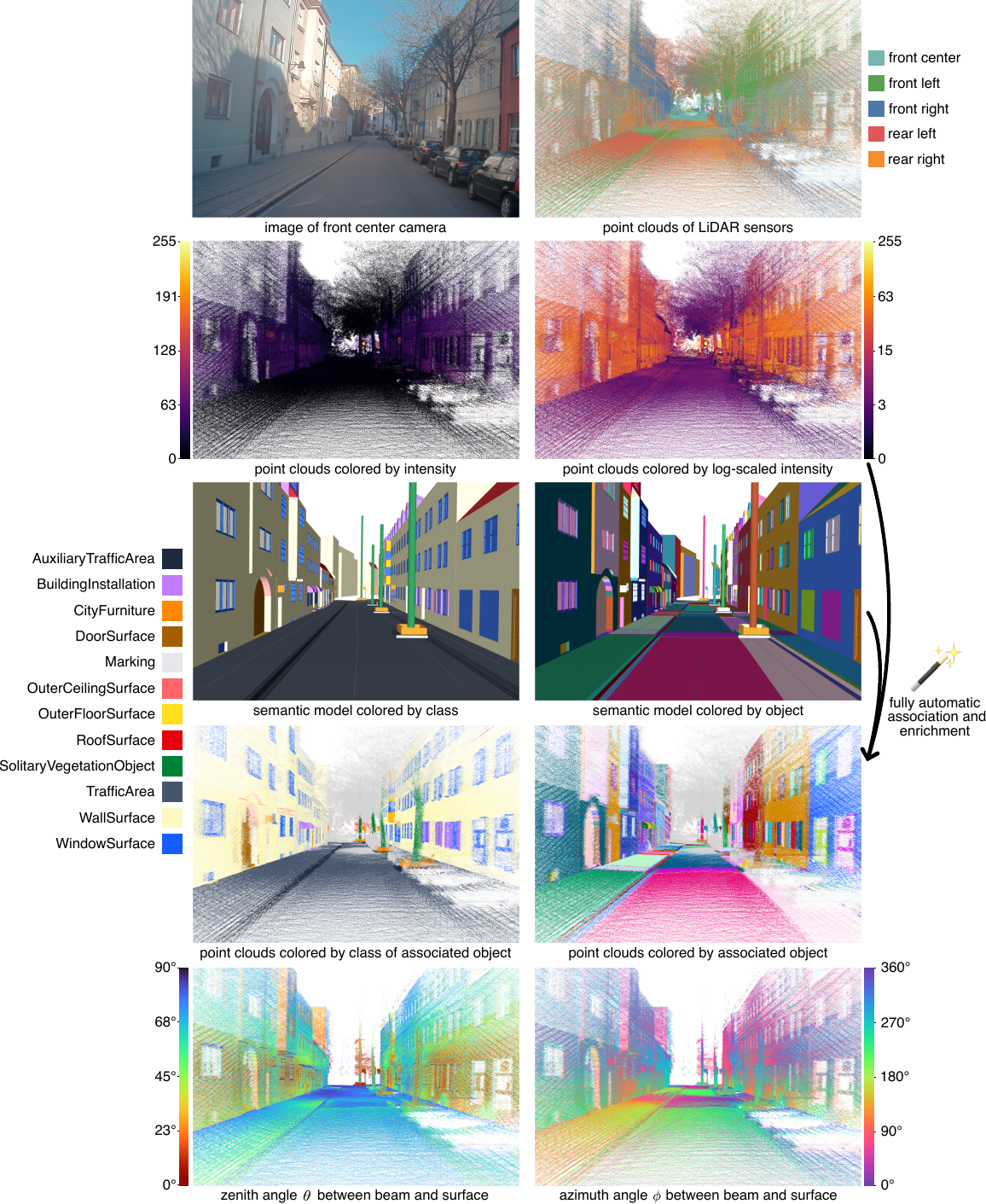}
    \caption{LiDAR point clouds from all sensor data acquisition campaigns were automatically associated with individual object surfaces of the semantic road space model and subsequently enriched using the proposed method. For visualization, only first-campaign point clouds are presented.}
    \label{fig:street-level-comparisons}
\end{widefigure}

\FloatBarrier
\subsection{Statistical analysis of the factors influencing the measured intensity}
To statistically analyze the factors influencing the measured intensity values across the sensors and campaigns, we perform the \gls{t-SNE} method directly on point cloud subsets.
\gls{t-SNE} is a nonlinear dimensionality reduction technique that assigns each datapoint a position on a two- or three-dimensional map while preserving local similarity structures by placing similar points nearby \citep{vandermaatenVisualizingDataUsing2008}.
The resulting maps are commonly used to explore high-dimensional data, as they reveal clusters and patterns that are not apparent in the original high-dimensional space.

\begin{figure}[htb]
    \centering
    \includegraphics[width=\linewidth]{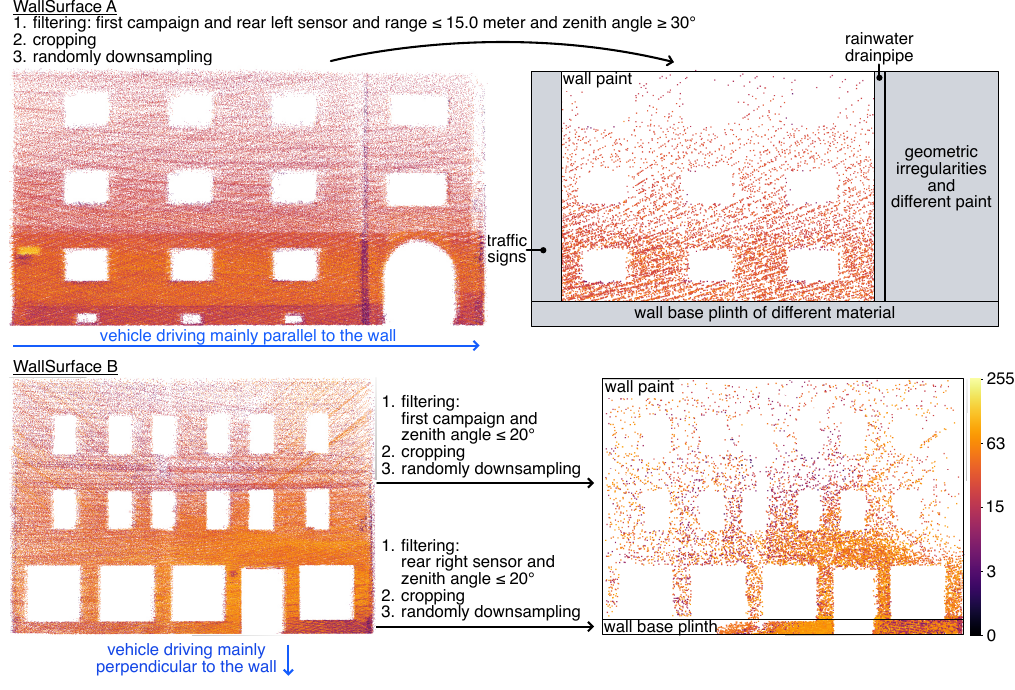}
    \caption{Selection of walls A and B for point-level analyses, as they were comprehensively captured by the sensors at varying ranges and angles. Additional cropping ensures that only points reflected from the same material are included.}
    \label{fig:wall-selection}
\end{figure}
\begin{figure}[htb]
    \centering
    \includegraphics[width=\linewidth]{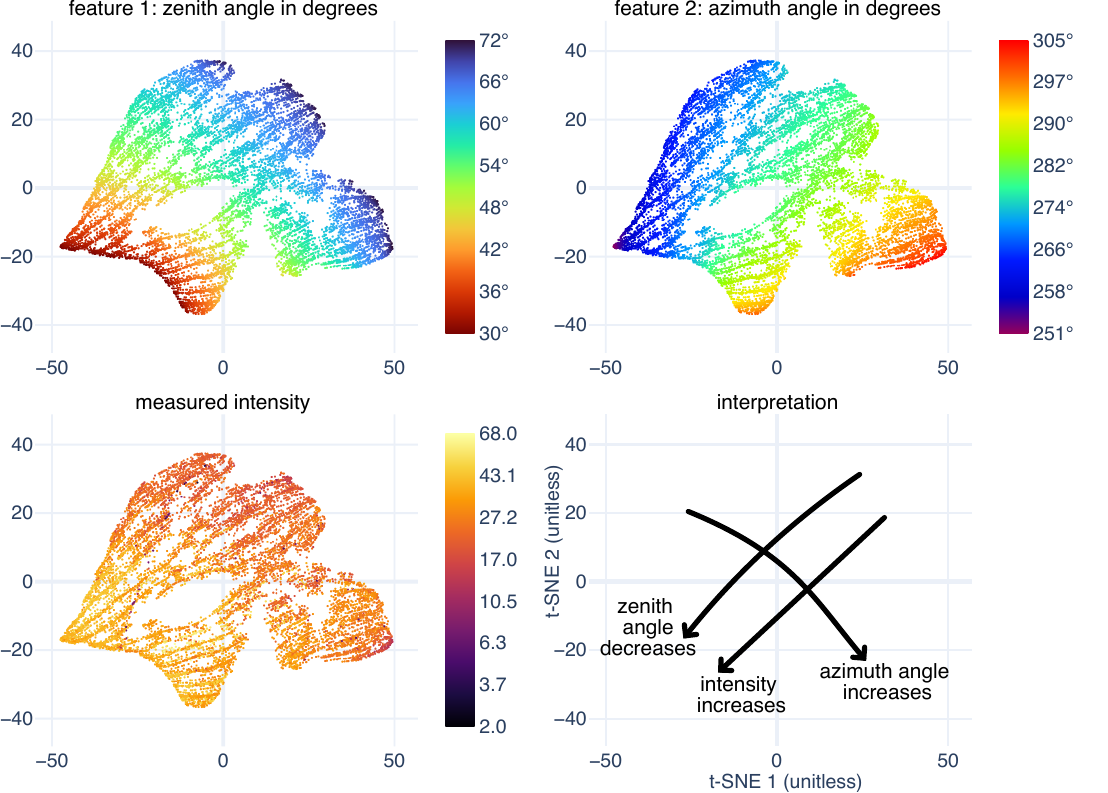}
    \caption{\gls{t-SNE} analysis at a perplexity of \num{200} for \num{10000} points reflected by the rear-left sensor on wall A. While the zenith angle  strongly influences the measured intensity, no such effect is observed for the azimuth angle.}
    \label{fig:tsne-wall-a}
\end{figure}

For this purpose, we select two larger walls, as shown in \autoref{fig:wall-selection}:
Since the vehicle moved parallel to wall A, it collected numerous beams within a 15-meter range at various zenith and azimuth angle combinations.
When moving orthogonally away from wall B, the vehicle's sensors captured measurements at varying ranges, but consistently at low zenith angles across all campaigns.
To ensure reflections originate only from a flat surface with a homogeneous material, the point cloud of wall A was manually cropped.
This step was required because elements, such as the rainwater drainpipe and wall base plinth, are not represented in the model, which leads to misassociations of points with low intensities for wall A.
To investigate the influence of zenith and azimuth angles on the measured intensity, only points from the first campaign and the rear-left sensor within a 15-meter range are considered, as the sensor exhibited comparatively consistent intensity measurements below \SI{15}{\meter} in the laboratory experiments.
Furthermore, the zenith angle is also restricted to $\geq 30^\circ$ to ensure a balanced distribution of zenith and azimuth angle pairs.
\autoref{fig:tsne-wall-a} presents the results of the \gls{t-SNE} analysis and reveals that lower zenith angles correspond to higher measured intensities, likely due to the combined effects of specular and diffuse reflection.
A correspondence of the intensity and the azimuth angle is not observable for wall A.
This may be attributed to isotropic material properties of the wall paint or to a negligible influence to the sensor operating at a wavelength of \SI{903}{\nano\meter}.

\begin{figure}[htb]
    \centering
    \includegraphics[width=.9\linewidth]{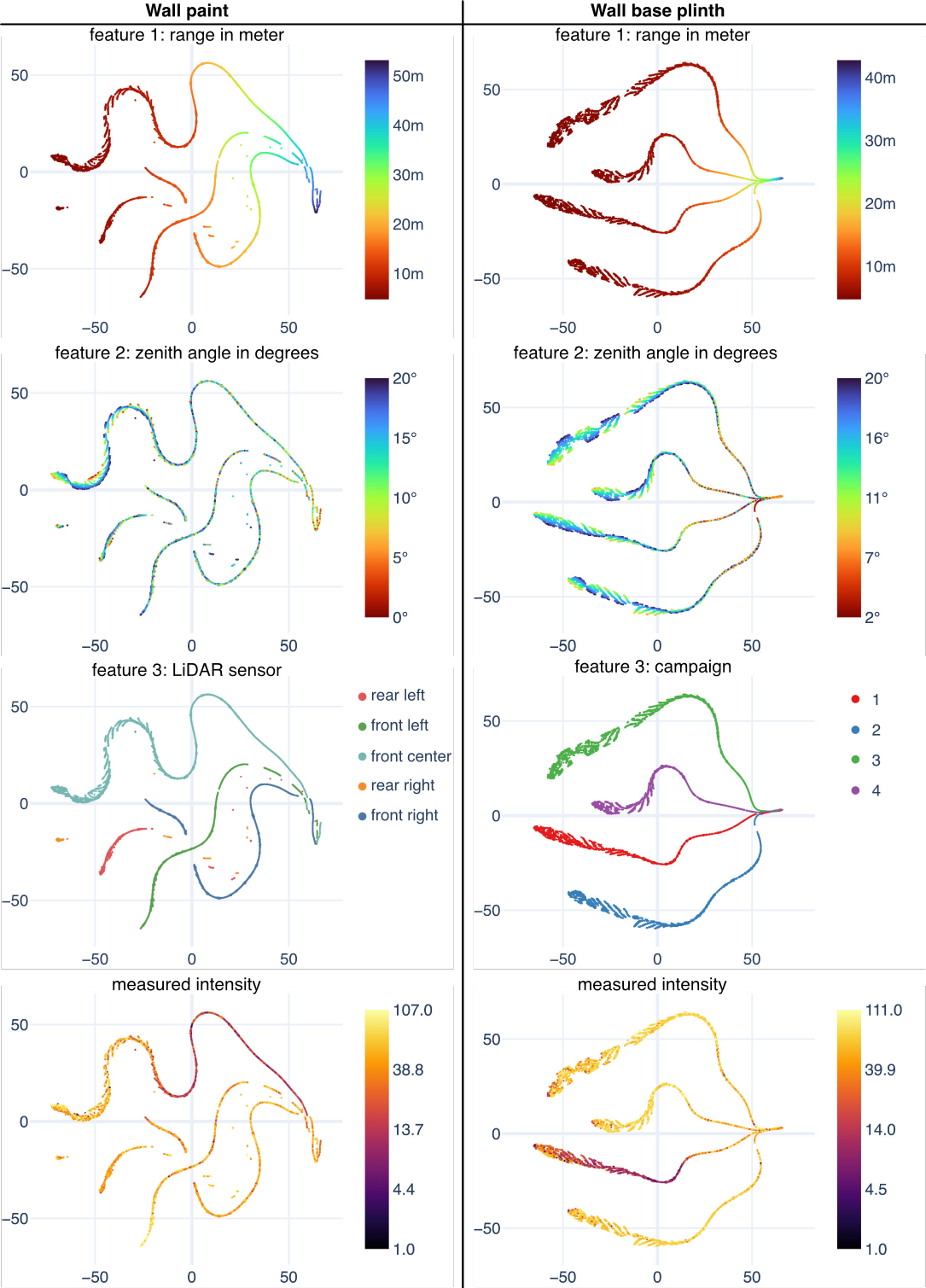}
    \caption{\gls{t-SNE} analysis for the wall paint (left) and wall base plinth (right) of wall B at perplexity 100 revealing that the measured intensities are influenced by the sensor, range, and campaign.}
    \label{fig:tsne-wall-b}
\end{figure}

To examine the effect of range on intensity measurements across all sensors, we select points reflected from the wall paint of wall B during the first campaign.
A maximum zenith angle restriction of \SI{20}{\degree} is applied to reduce confounding effects associated with angular incidence.
The outcome of the \gls{t-SNE} analysis using range, zenith angle, and the one-hot encoded sensor category as features is presented in the left diagram column of \autoref{fig:tsne-wall-b}.
Since \gls{t-SNE} primarily preserves local structure, distances between clusters in the resulting map lack interpretive value.
\gls{t-SNE} placed the points from the front-center sensor in an upper branch, which is highlighted in turquoise in the sub-diagram \emph{feature 3: LiDAR sensor}.
These points show a strong progression of the measured intensity from \num{13} to \num{100}, which corresponds to the trend of decreasing ranges from \num{52.5} to \SI{5.3}{\meter}.
This indicates a strong range dependence of the measured intensity for the front-center sensor.
However, such a strong effect is not observable for the front-left and front-right sensors, suggesting sensor-specific differences in response characteristics.

To analyze the influence of the campaign on measured intensity, we select points reflected from the wall base plinth of wall B, which were captured by the rear-right sensor across all campaigns.
The right column of \autoref{fig:tsne-wall-b} shows the \gls{t-SNE} embedding for the selected points.
Although campaigns 2-4, conducted within the same hour, exhibit similar intensity measurement characteristics, an apparent discrepancy is observed when compared to campaign 1, which took place five months earlier.
This discrepancy may be due to a wet surface during the first campaign, which could have absorbed more of the incident energy.
The higher surface wetness is likely attributable to the environmental conditions listed in \autoref{tab:environmental-conditions}, where drizzle on the previous day and nighttime temperatures down to \SI{-4}{\degreeCelsius} allowed the moisture to persist longer.

\FloatBarrier
\subsection{Deriving and comparing radiometric fingerprints of object surfaces}
\label{subsec:DerivingAndComparingRadiometricFingerprintsOfObjectSurfaces}
Since the range, zenith angle, sensor, and campaign impact the measured intensities, we derive radiometric fingerprints by grouping the intensity values associated with each object as follows:
\begin{align}
    \label{eq:radiometric-fingerprint-grouping}
    \mathcal{I}_{i,j}^{(c,s,o)} = \left\{ I_k \,\middle|\,
    \right. & r_k \in [r_i, r_{i+1}),\ \theta_k \in [\theta_j, \theta_{j+1}),\ \phi_k \in [0, 2\pi), \nonumber \\
            & \text{campaign}_k = c,\ \text{sensor}_k = s,\ \left. \text{object}_k = o \right\},
\end{align}
where $I_k$ denotes the measured intensity, $r_k$ the range, $\theta_k$ the zenith angle, and $\phi_k$ the azimuth angle of point $k$.
Given the comparatively consistent intensity measurements up to \SI{15}{\meter} and \SI{30}{\meter} in laboratory experiments, the range bin size is set to:
\begin{equation}
    \Delta r = r_{i+1} - r_i = \SI{15}{\meter}
\end{equation}
To achieve a trade-off between angular resolution and enough points per bin, the zenith angle domain is partitioned into the following intervals:
\begin{equation}
    [\theta_j, \theta_{j+1}) \in \left\{ [0^\circ, 20^\circ),\ [20^\circ, 40^\circ),\ [40^\circ, 60^\circ),\ [60^\circ, 90^\circ) \right\}
\end{equation}
For each bin $\mathcal{I}_{i,j}^{(c,s,o)}$, statistical descriptors of the intensity values $I_k$ are computed, such as the mean, standard deviation, median, first quartile, and third quartile.
The third quartile for each bin characterizes the upper intensity values while remaining robust to outliers, and is defined as:
\begin{equation}
    q_{i,j}^{(c,s,o)} = Q_3\left(\mathcal{I}_{i,j}^{(c,s,o)}\right)
\end{equation}
\autoref{fig:radiometric-fingerprint-plot} shows the third quartile as a function of the zenith angle bin for selected objects observed by the rear-left sensor within the range $[0\text{m}, 15\text{m})$ during the first campaign, as specified in \autoref{tab:radiometric-fingerprint-extraction-parameters-selected} in \ref{appendix:radiometric-fingerprint-extraction}.
\begin{widefigure}[htb]
    \centering
    \includegraphics[width=\linewidth]{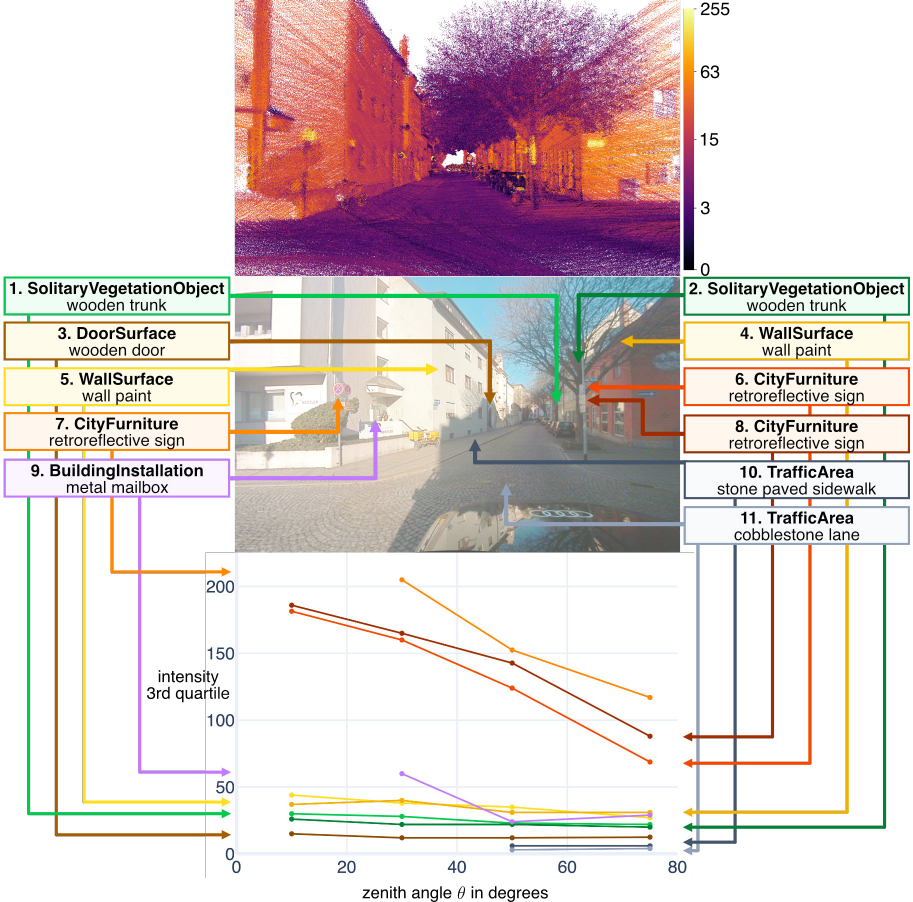}
    \caption{Radiometric fingerprints of individual objects observed by the rear-left sensor during the first campaign. The third intensity quartile of the associated points with $r<15\,\text{meter}$ reveals object-specific sensor responses caused by the different surface materials.}
    \label{fig:radiometric-fingerprint-plot}
\end{widefigure}
While the surfaces of individual objects reveal distinct sensor fingerprints, objects belonging to the same class exhibit similar intensity profiles.
For example, traffic signs show intensity values exceeding 150 within the zenith angle range $[0^\circ, 20^\circ)$, which can be attributed to their retroreflective surface coating.
The door surface and tree trunks are composed of wood and show lower intensity profiles compared to the wall paint for the complete zenith angle range.
The sidewalk and driving lane indicate an even lower reflectivity, whereby the surfaces are only observed within the zenith angle range $[40^\circ, 90^\circ)$ due to the sensor's mounting position and orientation.
The metal mailbox exhibits higher intensity values within the zenith angle range $[20^\circ, 40^\circ)$, while lower zenith angles are not observed due to occlusion by the postman.

\FloatBarrier
To evaluate the similarity between the radiometric fingerprints of object $o$ and $o'$, we compute an RMSE-based distance as follows:
\begin{equation}
    \text{dist}_{\text{Q3}}((c,s,o), (c',s',o')) =
    \sqrt{ \frac{1}{J} \sum_{j=1}^{J}
        \left( q_{i,j}^{(c,s,o)} - q_{i,j}^{(c',s',o')} \right)^2 },
\end{equation}
where the range bin index is fixed to $i=0$, corresponding to the range $[0\text{m}, 15\text{m})$.
\autoref{tab:individual-radiometric-fingerprint-distances} lists the pairwise distances for the selected objects from \autoref{fig:radiometric-fingerprint-plot} with a complete zenith angle coverage.

\begin{table}[htb]
    \footnotesize
    \centering
    \caption{RMSE-based distance $\text{dist}_{\text{Q3}}$ quantifying differences between the radiometric fingerprints of only those objects in \autoref{fig:radiometric-fingerprint-plot} that were fully covered by the sensor throughout all zenith-angle intervals $[\theta_j, \theta_{j+1})$.}
    \label{tab:individual-radiometric-fingerprint-distances}
    \begin{tabular}{@{} lrrrrrrrr @{}}
\toprule
\multicolumn{1}{c}{} & \multicolumn{1}{c}{1.} & \multicolumn{1}{c}{2.} & \multicolumn{1}{c}{3.} & \multicolumn{1}{c}{4.} & \multicolumn{1}{c}{5.} & \multicolumn{1}{c}{6.} & \multicolumn{1}{c}{8.} \\
\midrule
1. SolitaryVegetationObject &  0.0 & \cellcolor[rgb]{0.61,0.781,0.696} 3.8 & \cellcolor[rgb]{0.635,0.828,0.72} 13.2 & \cellcolor[rgb]{0.624,0.807,0.709} 9.2 & \cellcolor[rgb]{0.629,0.816,0.714} 10.8 & \cellcolor[rgb]{0.987,0.803,0.72} 114.9 & \cellcolor[rgb]{0.968,0.741,0.691} 124.3 \\
2. SolitaryVegetationObject &   &  0.0 & \cellcolor[rgb]{0.626,0.81,0.711} 9.7 & \cellcolor[rgb]{0.634,0.825,0.719} 12.7 & \cellcolor[rgb]{0.638,0.834,0.723} 14.1 & \cellcolor[rgb]{0.984,0.783,0.711} 118.3 & \cellcolor[rgb]{0.96,0.722,0.683} 127.7 \\
3. DoorSurface &   &   &  0.0 & \cellcolor[rgb]{0.695,0.865,0.739} 22.2 & \cellcolor[rgb]{0.709,0.871,0.742} 23.8 & \cellcolor[rgb]{0.958,0.718,0.681} 127.8 & \cellcolor[rgb]{0.923,0.662,0.661} 137.3 \\
4. WallSurface &   &   &   &  0.0 & \cellcolor[rgb]{0.611,0.784,0.697} 4.6 & \cellcolor[rgb]{0.995,0.863,0.748} 106.5 & \cellcolor[rgb]{0.987,0.799,0.718} 115.7 \\
5. WallSurface &   &   &   &   &  0.0 & \cellcolor[rgb]{0.997,0.878,0.756} 104.2 & \cellcolor[rgb]{0.989,0.815,0.725} 113.6 \\
6. CityFurniture &   &   &   &   &   &  0.0 & \cellcolor[rgb]{0.637,0.831,0.722} 13.9 \\
8. CityFurniture &   &   &   &   &   &   &  0.0 \\
\bottomrule
\end{tabular}
\end{table}

\FloatBarrier
\subsection{Semantics-based comparison of radiometric fingerprints}
\label{subsec:SemanticBasedComparisonOfRadiometricFingerprints}
To quantify the similarity of radiometric fingerprints between two sets of objects, $\mathcal{A}$ and $\mathcal{B}$, we compute the mean Q3-based distance over all object pairs with full coverage:
\begin{equation}
    \overline{d}_{\text{Q3}}(\mathcal{A}, \mathcal{B}) =
    \frac{1}{|\mathcal{P}|} \sum_{((c,s,o), (c',s',o')) \in \mathcal{P}}
    \text{dist}_{\text{Q3}}((c,s,o), (c',s',o')),
\end{equation}
where
\begin{equation}
    \mathcal{P} = \left\{ ((c,s,o), (c',s',o')) \in \mathcal{A} \times \mathcal{B} \;\middle|\; o \ne o' \right\}.
\end{equation}
\autoref{tab:overall-class-radiometric-fingerprint-distances} lists the mean distance $\overline{d}_{\text{Q3}}(\mathcal{A}, \mathcal{B})$ between each CityGML class, for which radiometric fingerprints were derived for the front-center sensor during the first campaign, as specified in \autoref{tab:radiometric-fingerprint-extraction-parameters-all} in \ref{appendix:radiometric-fingerprint-extraction}.
\begin{table}[htb]
    \footnotesize
    \centering
    \addtolength{\tabcolsep}{-0.4em}
    \caption{Mean distance $\overline{d}_{\text{Q3}}$ between the radiometric fingerprints of objects grouped by CityGML class indicating class-specific material composition patterns.}
    \label{tab:overall-class-radiometric-fingerprint-distances}
    \begin{tabular}{@{} r l rrrrrrrrrrr @{}}
\toprule
 & CityGML class & count &1. & 2. & 3. & 4. & 5. & 6. & 7. & 8. & 9. \\
\midrule
1. & AuxiliaryTrafficArea & 6 & \cellcolor[rgb]{0.63,0.819,0.715} 7.5 & \cellcolor[rgb]{0.884,0.951,0.781} 33.0 & \cellcolor[rgb]{0.92,0.659,0.661} 91.9 & \cellcolor[rgb]{0.887,0.952,0.784} 33.3 & \cellcolor[rgb]{0.893,0.955,0.788} 34.1 & \cellcolor[rgb]{0.784,0.907,0.758} 22.4 & \cellcolor[rgb]{0.695,0.865,0.739} 14.7 & \cellcolor[rgb]{0.94,0.975,0.818} 40.0 & \cellcolor[rgb]{0.868,0.944,0.771} 31.2 \\
2. & BuildingInstallation & 84 &   & \cellcolor[rgb]{0.871,0.945,0.773} 31.4 & \cellcolor[rgb]{0.986,0.795,0.716} 77.7 & \cellcolor[rgb]{0.881,0.949,0.779} 32.5 & \cellcolor[rgb]{0.815,0.921,0.761} 25.7 & \cellcolor[rgb]{0.815,0.921,0.761} 25.7 & \cellcolor[rgb]{0.903,0.959,0.794} 35.2 & \cellcolor[rgb]{0.896,0.956,0.79} 34.4 & \cellcolor[rgb]{0.868,0.944,0.771} 31.1 \\
3. & CityFurniture & 193 &   &   & \cellcolor[rgb]{0.985,0.787,0.712} 78.5 & \cellcolor[rgb]{0.986,0.795,0.716} 77.4 & \cellcolor[rgb]{0.991,0.831,0.733} 73.9 & \cellcolor[rgb]{0.984,0.783,0.711} 78.8 & \cellcolor[rgb]{0.908,0.647,0.661} 93.5 & \cellcolor[rgb]{0.989,0.815,0.725} 75.5 & \cellcolor[rgb]{0.984,0.779,0.709} 79.1 \\
4. & DoorSurface & 37 &   &   &   & \cellcolor[rgb]{0.887,0.952,0.784} 33.3 & \cellcolor[rgb]{0.835,0.929,0.763} 27.4 & \cellcolor[rgb]{0.831,0.927,0.763} 27.0 & \cellcolor[rgb]{0.915,0.964,0.802} 36.8 & \cellcolor[rgb]{0.893,0.955,0.788} 34.3 & \cellcolor[rgb]{0.878,0.948,0.777} 32.3 \\
5. & OuterFloorSurface & 1 &   &   &   &   &   & \cellcolor[rgb]{0.718,0.876,0.745} 16.5 & \cellcolor[rgb]{0.884,0.951,0.781} 33.1 & \cellcolor[rgb]{0.823,0.924,0.762} 26.2 & \cellcolor[rgb]{0.811,0.919,0.761} 25.0 \\
6. & SolitaryVegetationObject & 39 &   &   &   &   &   & \cellcolor[rgb]{0.709,0.871,0.742} 15.6 & \cellcolor[rgb]{0.811,0.919,0.761} 25.3 & \cellcolor[rgb]{0.854,0.938,0.766} 29.4 & \cellcolor[rgb]{0.811,0.919,0.761} 25.0 \\
7. & TrafficArea & 2 &   &   &   &   &   &   & \cellcolor[rgb]{0.768,0.9,0.756} 21.1 & \cellcolor[rgb]{0.952,0.98,0.834} 42.0 & \cellcolor[rgb]{0.884,0.951,0.781} 33.0 \\
8. & WallSurface & 195 &   &   &   &   &   &   &   & \cellcolor[rgb]{0.89,0.953,0.786} 33.7 & \cellcolor[rgb]{0.893,0.955,0.788} 34.2 \\
9. & WindowSurface & 206 &   &   &   &   &   &   &   &   & \cellcolor[rgb]{0.865,0.942,0.769} 30.7 \\
\bottomrule
\end{tabular}
\end{table}
The results reveal systematic differences between the radiometric fingerprints of CityFurniture objects, for example, and those of other classes.
Furthermore, the CityFurniture class shows a significantly higher mean intra-class distance of $78.5$, suggesting considerable variation in material composition.
This is confirmed by grouping only the CityFurniture objects by their function and computing mean distances accordingly, as listed in \autoref{tab:city-furniture-radiometric-fingerprint-distances}.
\begin{table}[htb]
    \footnotesize
    \centering
    \addtolength{\tabcolsep}{-0.45em}
    \caption{Mean distance $\overline{d}_{\text{Q3}}$ between radiometric fingerprints of CityFurniture objects grouped by function demonstrating function-specific material patterns.}
    \label{tab:city-furniture-radiometric-fingerprint-distances}
    \begin{tabular}{@{} r l rrrrrrrrrrrrrrrr @{}}
\toprule
 & function of object & count &1. & 2. & 3. & 4. & 5. & 6. & 7. & 8. & 9. & 10. & 11. & 12. & 13. & 14. \\
\midrule
1. & bench & 5 & \cellcolor[rgb]{0.811,0.919,0.761} 25 & \cellcolor[rgb]{0.835,0.929,0.763} 28 & \cellcolor[rgb]{0.843,0.933,0.764} 28 & \cellcolor[rgb]{0.937,0.974,0.816} 40 & \cellcolor[rgb]{1.0,0.988,0.879} 53 & \cellcolor[rgb]{0.764,0.898,0.756} 20 & \cellcolor[rgb]{0.997,0.871,0.751} 70 & \cellcolor[rgb]{0.994,0.998,0.892} 49 & \cellcolor[rgb]{1.0,0.999,0.898} 50 & \cellcolor[rgb]{0.887,0.627,0.66} 96 & \cellcolor[rgb]{0.859,0.6,0.66} 123 & \cellcolor[rgb]{0.859,0.6,0.66} 119 & \cellcolor[rgb]{0.997,0.878,0.756} 70 & \cellcolor[rgb]{0.78,0.905,0.757} 22 \\
2. & controller box & 4 &   & \cellcolor[rgb]{0.827,0.926,0.763} 27 & \cellcolor[rgb]{0.912,0.963,0.8} 37 & \cellcolor[rgb]{0.9,0.957,0.792} 35 & \cellcolor[rgb]{0.998,0.953,0.821} 60 & \cellcolor[rgb]{0.871,0.945,0.773} 31 & \cellcolor[rgb]{0.999,0.961,0.834} 58 & \cellcolor[rgb]{1.0,0.999,0.898} 50 & \cellcolor[rgb]{0.896,0.956,0.79} 35 & \cellcolor[rgb]{0.97,0.745,0.693} 83 & \cellcolor[rgb]{0.859,0.6,0.66} 106 & \cellcolor[rgb]{0.859,0.6,0.66} 102 & \cellcolor[rgb]{0.998,0.933,0.803} 62 & \cellcolor[rgb]{0.776,0.903,0.757} 22 \\
3. & fence & 3 &   &   & \cellcolor[rgb]{0.893,0.955,0.788} 34 & \cellcolor[rgb]{0.98,0.992,0.872} 47 & \cellcolor[rgb]{1.0,0.995,0.892} 51 & \cellcolor[rgb]{0.799,0.914,0.76} 24 & \cellcolor[rgb]{0.983,0.771,0.705} 80 & \cellcolor[rgb]{1.0,0.995,0.892} 51 & \cellcolor[rgb]{0.999,0.959,0.831} 58 & \cellcolor[rgb]{0.859,0.6,0.66} 105 & \cellcolor[rgb]{0.859,0.6,0.66} 132 & \cellcolor[rgb]{0.859,0.6,0.66} 129 & \cellcolor[rgb]{0.989,0.815,0.725} 76 & \cellcolor[rgb]{0.868,0.944,0.771} 31 \\
4. & other & 11 &   &   &   & \cellcolor[rgb]{0.982,0.993,0.876} 47 & \cellcolor[rgb]{0.997,0.902,0.777} 66 & \cellcolor[rgb]{0.952,0.98,0.834} 42 & \cellcolor[rgb]{0.998,0.933,0.803} 62 & \cellcolor[rgb]{0.999,0.961,0.834} 58 & \cellcolor[rgb]{0.945,0.977,0.824} 41 & \cellcolor[rgb]{0.972,0.748,0.695} 82 & \cellcolor[rgb]{0.859,0.6,0.66} 102 & \cellcolor[rgb]{0.868,0.609,0.66} 99 & \cellcolor[rgb]{0.997,0.884,0.761} 68 & \cellcolor[rgb]{0.893,0.955,0.788} 34 \\
5. & pole & 4 &   &   &   &   & \cellcolor[rgb]{0.983,0.775,0.707} 79 & \cellcolor[rgb]{0.982,0.993,0.876} 47 & \cellcolor[rgb]{0.94,0.681,0.664} 89 & \cellcolor[rgb]{0.998,0.911,0.784} 65 & \cellcolor[rgb]{0.99,0.823,0.729} 75 & \cellcolor[rgb]{0.859,0.6,0.66} 103 & \cellcolor[rgb]{0.859,0.6,0.66} 122 & \cellcolor[rgb]{0.859,0.6,0.66} 120 & \cellcolor[rgb]{0.972,0.748,0.695} 82 & \cellcolor[rgb]{0.999,0.978,0.863} 54 \\
6. & railing & 13 &   &   &   &   &   & \cellcolor[rgb]{0.666,0.851,0.732} 12 & \cellcolor[rgb]{0.986,0.795,0.716} 78 & \cellcolor[rgb]{0.987,0.995,0.882} 48 & \cellcolor[rgb]{0.999,0.967,0.844} 57 & \cellcolor[rgb]{0.859,0.6,0.66} 103 & \cellcolor[rgb]{0.859,0.6,0.66} 131 & \cellcolor[rgb]{0.859,0.6,0.66} 127 & \cellcolor[rgb]{0.992,0.839,0.736} 73 & \cellcolor[rgb]{0.776,0.903,0.757} 22 \\
7. & sign (private) & 11 &   &   &   &   &   &   & \cellcolor[rgb]{0.998,0.948,0.815} 60 & \cellcolor[rgb]{0.989,0.815,0.725} 76 & \cellcolor[rgb]{0.975,0.99,0.866} 46 & \cellcolor[rgb]{0.998,0.924,0.795} 63 & \cellcolor[rgb]{0.992,0.839,0.736} 73 & \cellcolor[rgb]{0.995,0.863,0.748} 71 & \cellcolor[rgb]{0.99,0.823,0.729} 75 & \cellcolor[rgb]{0.998,0.927,0.797} 63 \\
8. & street lamp & 12 &   &   &   &   &   &   &   & \cellcolor[rgb]{0.998,0.927,0.797} 63 & \cellcolor[rgb]{0.999,0.957,0.828} 59 & \cellcolor[rgb]{0.93,0.668,0.661} 91 & \cellcolor[rgb]{0.859,0.6,0.66} 110 & \cellcolor[rgb]{0.859,0.6,0.66} 108 & \cellcolor[rgb]{0.995,0.859,0.746} 71 & \cellcolor[rgb]{0.989,0.996,0.885} 48 \\
9. & traffic light & 1 &   &   &   &   &   &   &   &   &   & \cellcolor[rgb]{0.998,0.93,0.8} 63 & \cellcolor[rgb]{0.983,0.775,0.707} 80 & \cellcolor[rgb]{0.988,0.807,0.722} 76 & \cellcolor[rgb]{0.999,0.955,0.824} 59 & \cellcolor[rgb]{0.937,0.974,0.816} 40 \\
10. & traffic sign (guidance) & 10 &   &   &   &   &   &   &   &   &   & \cellcolor[rgb]{0.998,0.953,0.821} 59 & \cellcolor[rgb]{0.999,0.976,0.86} 55 & \cellcolor[rgb]{0.999,0.982,0.869} 54 & \cellcolor[rgb]{0.976,0.756,0.698} 82 & \cellcolor[rgb]{0.945,0.692,0.669} 88 \\
11. & traffic sign (regul.) & 37 &   &   &   &   &   &   &   &   &   &   & \cellcolor[rgb]{0.975,0.99,0.866} 46 & \cellcolor[rgb]{0.973,0.989,0.863} 46 & \cellcolor[rgb]{0.911,0.65,0.661} 93 & \cellcolor[rgb]{0.859,0.6,0.66} 113 \\
12. & traffic sign (suppl.) & 39 &   &   &   &   &   &   &   &   &   &   &   & \cellcolor[rgb]{0.975,0.99,0.866} 46 & \cellcolor[rgb]{0.92,0.659,0.661} 92 & \cellcolor[rgb]{0.859,0.6,0.66} 109 \\
13. & traffic sign pole & 40 &   &   &   &   &   &   &   &   &   &   &   &   & \cellcolor[rgb]{0.997,0.871,0.751} 70 & \cellcolor[rgb]{0.998,0.918,0.79} 64 \\
14. & wall & 3 &   &   &   &   &   &   &   &   &   &   &   &   &   & \cellcolor[rgb]{0.776,0.903,0.757} 22 \\
\bottomrule
\end{tabular}
\end{table}
For instance, both the concrete-wood composite benches and the plastered walls exhibit low intensity profiles.
Private signs, such as non-retroreflective advertising signs, yield lower intensity profiles compared to traffic signs, which are consistently coated with retroreflective material.

\subsection{Limitations}
For grouping LiDAR sensor measurements and enriching them with further semantic information, the association method presented relies on a semantic 3D road space model.
If an object exists in reality but is not represented in the model, the corresponding LiDAR reflections remain unassociated unless the object lies within the defined tolerance range of another modeled object.
For example, LiDAR points reflected from wall-mounted signs that are not represented in the model are erroneously associated with the wall due to spatial proximity.
Conversely, an object represented in the model but absent in reality does not lead to false associations, provided it does not occlude another object in its spatial proximity.

The evaluation and comparison of radiometric fingerprints were performed exclusively on objects exhibiting complete coverage of the zenith-angle bins within the specified range bin.
To determine the true reflectance values of object surfaces, prior radiometric correction and calibration (Level 3) are required.
This step enables the direct assessment of material characteristics and, by normalizing the acquisition geometry, allows all measurements to be combined without the need for bin-wise analysis.
If radiometric fingerprints still cannot be derived for a desired number of object surfaces, coverage can be increased by conducting additional campaigns from different directions over multiple epochs.

Since the semantic 3D road space model represents the tree trunk of SolitaryVegetationObjects with abstract cylindrical geometries, points reflected from branches and leaves in the tree crown remain unassociated, as shown in \autoref{fig:association-comparison} in \ref{appendix:automatic-association}.
While this enables the extraction of the radiometric fingerprint of the tree trunk, vegetation growth dynamics necessitate approximating the tree crown for each epoch \citep{hirtChangeDetectionUrban2021} or performing a pointwise leaf-wood classification \citep{wangLeWoSUniversalLeafwood2020}.
Because LiDAR beams commonly pass through glass and are reflected by interior room surfaces, the resulting returns are correctly not associated with the surface geometries of window objects.
Consequently, the point density on windows is lower when reflections occur only on the window frame while the curtains are open.

In conducted study, objects are assumed to consist of a homogeneous material, which is largely valid for the objects in the semantic 3D road space model.
To address this, object surfaces in the \gls{LOD}3 semantic road space model are to be further subdivided by material and corresponding CityGML appearance definitions incorporated.

The association process of LiDAR measurements with object surfaces from the model is subject to various sources of uncertainty.
For \gls{MLS} systems, these uncertainties are commonly classified into four main categories: instrumental errors, atmospheric errors, object and geometry-related errors, trajectory estimation errors \citep{xuPL4UAutomatedPlanebased2025}.
Each error source introduces both random and systematic errors into the final point cloud.
Constructing a complete error model would require determining all relevant mathematical and physical correlations, many of which are still unknown \citep{holstChallengesPresentFields2016,kerekesDeterminingVarianceCovarianceMatrices2021,schmitzEmpiricalDeterminationCorrelations2021}.
For example, laboratory measurements indicate that the positional accuracy of the VLP-16 LiDAR sensor decreases with increasing distance and decreasing reflectivity of the target surface beyond the reported absolute accuracy in the specification.
Due to the many possible influencing factors and their interdependencies, accuracy is typically assessed using the final point cloud.
Accordingly, the point cloud submaps obtained for each campaign were evaluated against the reference point cloud listed in \autoref{tab:fitness_rmse_scores}.
While the reference \gls{MLS} point cloud achieves relative accuracy in the millimeter range \citep{haigermoserRoadTrackIrregularities2015}, the derived road network model and the \gls{LOD}3 building models are affected by geometric abstractions and manual modeling inaccuracies.
In contrast, the \gls{LOD}2 building models are reconstructed from different surveys by the State Mapping Agency and offer an absolute accuracy of up to \SI{3}{\centi\meter}.
Despite combining semantic models from multiple surveying campaigns using different sensor equipment and facing various sources of uncertainty, the study demonstrates that the association process performs sufficiently well to enable the extraction and comparison of radiometric fingerprints for even smaller objects such as the traffic signs or mailboxes.

\section{Conclusion and outlook}
In this work, we demonstrate how individual LiDAR sensor measurements acquired by an automated vehicle across multiple campaigns can be associated with object surfaces of a detailed semantic 3D city model.
Using our proposed ray-casting-based association method, we automatically link \num{314.4} million points to the \num{15816} thematic objects of the semantic 3D road space models in CityGML 3.0 that represent the study area.
This approach has the potential to eliminate or substantially reduce the tedious manual labeling of point clouds for static environment objects, which often hampers learning-based segmentation and reconstruction tasks.
A semantic 3D model becomes particularly valuable for auto-labelling repeated scans of the same environment from different positions and orientations, under varying atmospheric and wetness conditions, and with diverse sensor hardware and configurations.
We expect that the efficient generation of closer real-world distributed training data will improve learning-based point cloud segmentation and 3D object reconstruction in general.
Moreover, the fine-grained subdivision of objects in the semantic 3D model enables the flexible derivation of class lists tailored to the requirements of point cloud semantic segmentation tasks, which can be subsequently modified and reassociated as needed.
In the context of urban digital twins and their applications, the updating of semantic 3D city models remains a key challenge \citep{lehtolaDigitalTwinCity2022}.
For this purpose, the association method can be leveraged to detect deviations between the semantic 3D model and repeated scans of the road space, acquired either by automated vehicles or by LiDAR sensors mounted on buses or garbage trucks.

Despite radiometric inaccuracies identified in laboratory measurements of the deployed LiDAR sensor, our study demonstrates that material-specific reflection patterns of individual object surfaces in the road space can be detected using an automated vehicle equipped with mid-tier LiDAR sensors.
Our statistical analysis shows that surfaces composed of the same material exhibit similar radiometric fingerprints, whereas surfaces of different materials display distinct intensity profiles, resulting in larger fingerprint distances.
For example, the fingerprints of private signs and traffic signs have a great average distance of approximately 70 out of 255 (unitless), which is caused by the retroreflective coating of the traffic signs.
In the future, radiometric fingerprints could serve as input for data-driven LiDAR sensor models, whereby standard-based semantic 3D models facilitate their transfer and application across different locations.
In addition, the proposed fingerprints may be utilized to estimate pre-scanned materials from material libraries and incorporate them into semantic 3D models.
Accordingly, further fingerprints of object surfaces could be obtained through additional sensor modalities, such as RGB, thermal, and spectral cameras, as well as radar sensors.
To support this, we plan to expand the introduced 3DSensorDB and generalize it for further use cases.
Extending CityGML with a conceptual data model for physical material properties would enhance urban digital twin applications, including building energy estimation, microclimate simulation, and sensor simulation.


\section*{CRediT authorship contribution statement}
\textbf{Benedikt Schwab}: Conceptualization, Methodology, Software, Validation, Investigation, Resources, Data curation, Writing - Original Draft, Writing - Review \& Editing, Visualization. \textbf{Thomas H. Kolbe}: Writing - Review \& Editing, Supervision, Project administration, Funding acquisition.

\section*{Declaration of competing interest}
The authors declare that they have no known competing financial interests or personal relationships that could have appeared to influence the work reported in this paper.

\section*{Acknowledgements}
We thank Lutz Morich and Wolfram Remlinger for initiating the research projects SAVe and SAVeNoW at Audi\,AG and for promoting technical progress.
Moreover, we are grateful to Kevin Bondzio and Sebastian Dorn from Audi's \gls{A2D2} team for their support and the opportunity to carry out an additional sensor data acquisition campaigns in the \gls{LOD}3 area.
Finally, we thank Sophie Haas Goschenhofer and Olaf Wysocki for their support in modeling the \gls{LOD}3 building models.

\section*{Data and code availability statement}
The introduced 3DSensorDB, the semantic 3D road space models in CityGML 3.0, the modelling guideline, and the associated point clouds are released at \url{https://github.com/tum-gis/sensordb}.
The converter from \gls{A2D2} datasets into the bag format of \gls{ROS} is released at \url{https://github.com/tum-gis/a2d2_ros_preparer} along with the Cartographer configurations at \url{https://github.com/tum-gis/cartographer_audi_a2d2}.


\appendix
\newpage
\section{Sensor specification}
\begin{table}[htb]
    \footnotesize
    \centering
    \caption{Specifications of the VLP-16 LiDAR sensor by \cite{velodynelidarinc.VLP16UserManual2022}.}
    \label{tab:velodyne-vlp-16-specification}
    \begin{tabular}{@{} ll @{}}
        \toprule
        Parameter                 & Specification                                                  \\
        \midrule
        Channels                  & \num{16}                                                       \\
        Field of View: Vertical   & \SI{30}{\degree} (\SI{-15}{\degree} to \SI{+15}{\degree})      \\
        Measurement range         & \SI{100}{\meter}                                               \\
        Field of View: Horizontal & \SI{360}{\degree}                                              \\
        Rotation Rate             & \SIrange{5}{20}{\hertz} (\SI{10}{\hertz} used for experiments) \\
        Range accuracy            & up to ±\SI{3}{\centi\meter} (typical)                          \\
        Rate                      & up to ~\num{300000} points/second                              \\
        Wavelength                & \SI{903}{\nano\meter}                                          \\
        \bottomrule
    \end{tabular}
\end{table}

\newpage
\section{Semantic road space model}
\begin{widefigure}[htb]
    \centering
    \includegraphics[width=\linewidth]{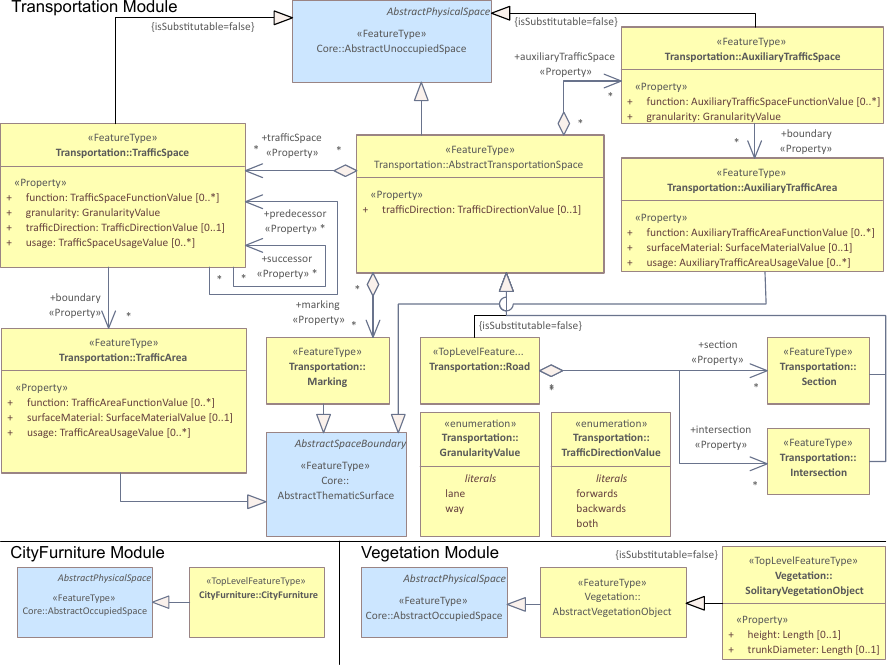}
    \caption{CityGML 3.0 classes utilized to represent the road network and roadside objects.}
    \label{fig:road-network-uml}
\end{widefigure}

\begin{widefigure}[htb]
    \raggedright
    \includegraphics[width=.94\linewidth]{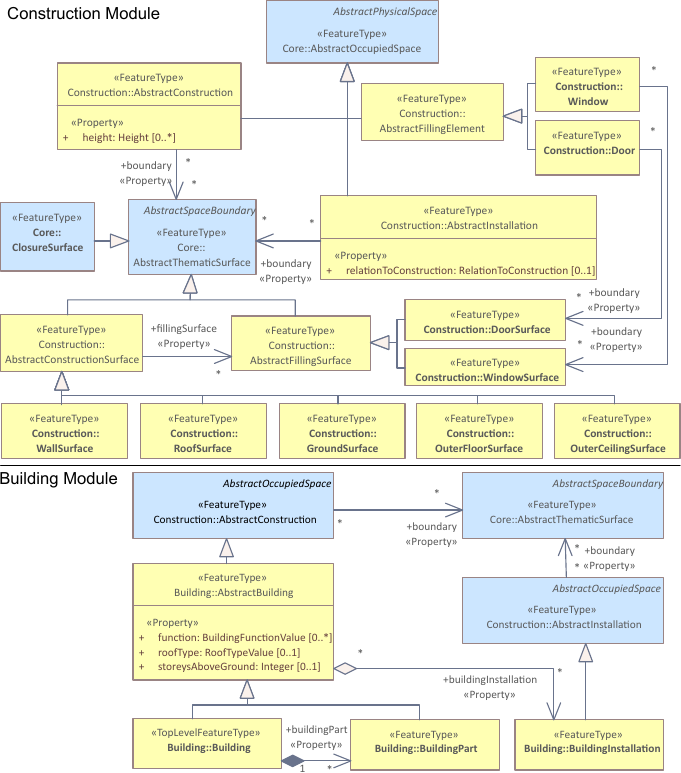}
    \caption{CityGML 3.0 classes of the Construction and Building module used to represent the \gls{LOD}2 and 3 building models.}
    \label{fig:building-model-uml}
\end{widefigure}

\FloatBarrier
\section{Automatic association}
\label{appendix:automatic-association}
\begin{table}[htb]
    \footnotesize
    \centering
    \caption{Parameter values utilized for the beam-surface association process.}
    \label{tab:automatic-association-parameters}
    \begin{tabular}{@{} ll @{}}
        \toprule
        Parameter                                    & Value                  \\
        \midrule
        Line segment length $\ell$                   & $\SI{1}{\meter}$       \\
        Spherocylinder radius $\rho$                 & $\SI{5}{\centi\meter}$ \\
        Max.\ object surface associations per  point & 1                      \\
        \bottomrule
    \end{tabular}
\end{table}
\begin{widefigure}[htb]
    \centering
    \includegraphics[width=.62\linewidth]{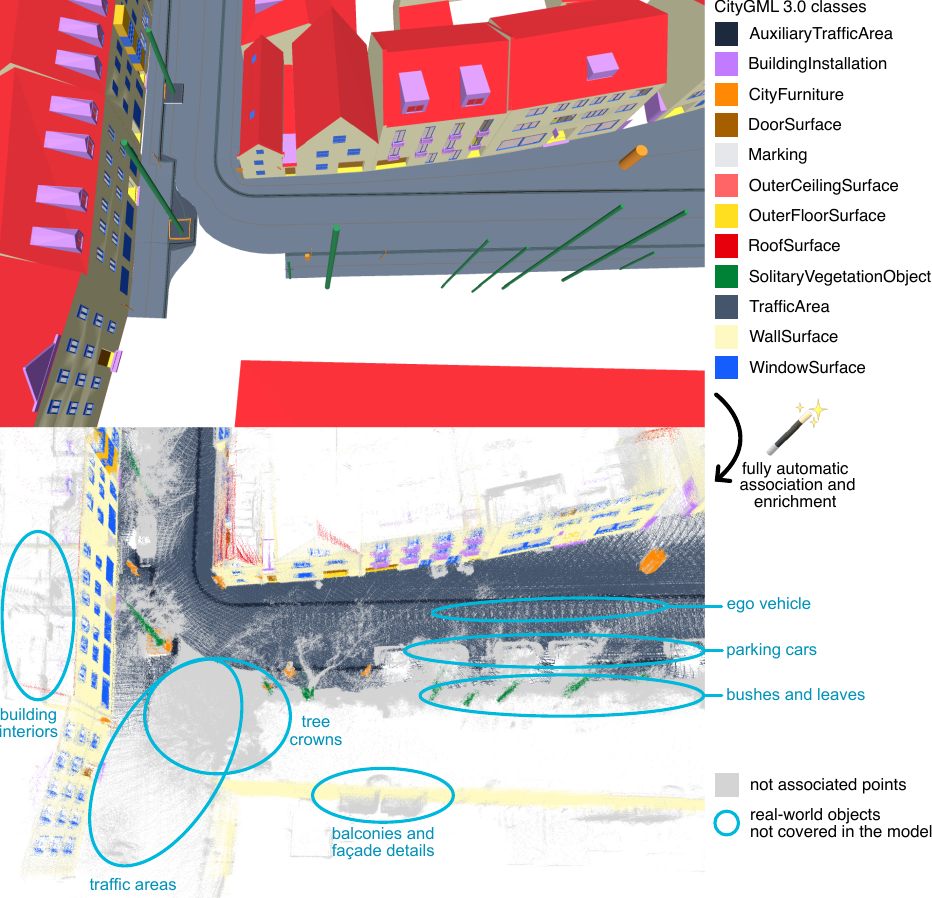}
    \caption{Semantic road space model (top) and automatically associated point cloud from campaign 1 (bottom). While \SI{71.7}{\percent} of the points were successfully associated, the remaining unassociated points are primarily due to objects or object details that are not represented in the semantic 3D model.}
    \label{fig:association-comparison}
\end{widefigure}

\newpage
\FloatBarrier
\section{Radiometric fingerprint extraction}
\label{appendix:radiometric-fingerprint-extraction}
\begin{table}[htb]
    \footnotesize
    \centering
    \caption{Filters applied according to \autoref{eq:radiometric-fingerprint-grouping} for extracting radiometric fingerprints, as discussed in Section \ref{subsec:DerivingAndComparingRadiometricFingerprintsOfObjectSurfaces} and illustrated in \autoref{fig:radiometric-fingerprint-plot} and \autoref{tab:individual-radiometric-fingerprint-distances}.}
    \label{tab:radiometric-fingerprint-extraction-parameters-selected}
    \begin{tabular}{@{} ll @{}}
        \toprule
        Parameter               & Value                                                                                                       \\
        \midrule
        Range $r_k$             & $[0\text{m}, 15\text{m})$                                                                                   \\
        Zenith angle $\theta_k$ & $\left\{ [0^\circ, 20^\circ),\ [20^\circ, 40^\circ),\ [40^\circ, 60^\circ),\ [60^\circ, 90^\circ) \right\}$ \\
        Azimuth angle $\phi_k$  & $[0,2\pi)$                                                                                                  \\
        $\text{Campaign}_k$     & campaign 1                                                                                                  \\
        $\text{Sensor}_k$       & rear-left sensor                                                                                            \\
        $\text{Object}_k$       & selected objects highlighted in \autoref{fig:radiometric-fingerprint-plot}                                  \\
        \bottomrule
    \end{tabular}
\end{table}

\begin{table}[htb]
    \footnotesize
    \centering
    \caption{Filters applied according to \autoref{eq:radiometric-fingerprint-grouping} for extracting radiometric fingerprints, as discussed in Section \ref{subsec:SemanticBasedComparisonOfRadiometricFingerprints} and listed in \autoref{tab:overall-class-radiometric-fingerprint-distances} and \ref{tab:city-furniture-radiometric-fingerprint-distances}.}
    \label{tab:radiometric-fingerprint-extraction-parameters-all}
    \begin{tabular}{@{} ll @{}}
        \toprule
        Parameter               & Value                                                                                                       \\
        \midrule
        Range $r_k$             & $[0\text{m}, 15\text{m})$                                                                                   \\
        Zenith angle $\theta_k$ & $\left\{ [0^\circ, 20^\circ),\ [20^\circ, 40^\circ),\ [40^\circ, 60^\circ),\ [60^\circ, 90^\circ) \right\}$ \\
        Azimuth angle $\phi_k$  & $[0,2\pi)$                                                                                                  \\
        $\text{Campaign}_k$     & campaign 1                                                                                                  \\
        $\text{Sensor}_k$       & front-center sensor                                                                                         \\
        $\text{Object}_k$       & --                                                                                                          \\
        \bottomrule
    \end{tabular}
\end{table}

\FloatBarrier




\bibliographystyle{elsarticle-harv}
\bibliography{cas-refs}
\end{document}